\documentclass[10pt,journal,compsoc]{IEEEtran}

\usepackage{times}
\usepackage{epsfig}
\usepackage{graphicx}
\usepackage{amsmath}
\usepackage{amssymb}
\usepackage{color}
\usepackage{subfig}
\usepackage{framed}
\usepackage[dvipsnames]{xcolor}
\usepackage{enumitem}
\usepackage[normalem]{ulem} %%% added for crossing out text

\usepackage{multirow}
\usepackage{tcolorbox}
\usepackage{ifthen} %added by AF
\usepackage{makecell} %added by AF
\definecolor{EPIC-BLUE}{HTML}{00B6D6} %added by AF
\definecolor{EPIC-RED}{HTML} {ED323E} %added by AF
\usepackage{multicol} % added by HD
\usepackage{booktabs} % added by HD

\definecolor{EPIC-COLOR}{HTML}{ED323E}
\definecolor{EPIC-COLOR2}{HTML}{00D6D6}

\newcommand{\EPIC}{{EPIC-KITCHENS}}

\newcommand{\fps}{\textit{fps}}

\newcommand{\added}[1]{\textcolor{black}{#1}}
\newcommand{\revision}[1]{\textcolor{black}{#1}}

\usepackage[hidelinks]{hyperref}

\DeclareMathOperator{\argmax}{\arg\!\max}
\DeclareMathOperator{\argmin}{\arg\!\min}
\DeclareMathOperator*{\tvl}{TV-L_1\!}

\newcommand*{\jhat}{\hat{\jmath}}

\ifCLASSOPTIONcompsoc
  \usepackage[nocompress]{cite}
\else
  \usepackage{cite}
\fi

\ifCLASSINFOpdf
\else
\fi

\begin{document}
\title{The \EPIC{} Dataset:\\ Collection, Challenges and Baselines}

% Return this one - Dima
\author{Dima~Damen, %~\IEEEmembership{Member,~IEEE,}
Hazel Doughty,
Giovanni Maria Farinella,
Sanja Fidler,
Antonino Furnari,\\
Evangelos Kazakos,
Davide Moltisanti,
Jonathan Munro,
Toby Perrett,
Will Price,
Michael Wray

\IEEEcompsocitemizethanks{
\IEEEcompsocthanksitem Authors list sorted alphabetically
\IEEEcompsocthanksitem D. Damen, H. Doughty, E. Kazakos, D. Moltisanti, J. Munro, T. Perrett, W. Price and M. Wray are with the Department
of Computer Science, University of Bristol, UK. Email: $<$firstname$>$.$<$lastname$>$@bristol.ac.uk
\IEEEcompsocthanksitem S. Fidler is with the University of Toronto, Vector Institute and NVIDIA, Canada. Email: fidler@cs.toronto.edu
\IEEEcompsocthanksitem G. M. Farinella and A. Furnari are with the University of Catania, Italy. Email: $<$gfarinella,furnari $>$@dmi.unict.it
\IEEEcompsocthanksitem The authors thank all 32 participants in the dataset collection.
\IEEEcompsocthanksitem Research at the University of Bristol is supported by Engineering \& Physical Sciences Research Council (EPSRC) Doctoral Training Programme, project GLANCE (EP/N013964/1), project LOCATE (EP/N033779/1) and fellowship UMPIRE (EP/T004991/1). Annotations have been sponsored by a charitable donation from Nokia Technologies and the University of Bristol's Jean Golding Institute.
\IEEEcompsocthanksitem Research at the University of Catania is sponsored by Piano della Ricerca 2016-2018 – linea di Intervento~2 of DMI.}
}


\markboth{April 2020}%
{Damen \MakeLowercase{\textit{et al.}}: EPIC-KITCHENS Dataset}



\IEEEtitleabstractindextext{%
\begin{abstract}
Since its introduction in 2018, \EPIC{} has attracted attention as the largest egocentric video benchmark, offering a unique viewpoint on people's interaction with objects, their attention, and even intention.
In this paper, we detail how this large-scale dataset was captured by 32 participants in their native kitchen environments, and densely annotated with actions and object interactions. Our videos depict \textbf{nonscripted} daily activities, as recording is started every time a participant entered their kitchen. Recording took place in 4 countries by participants belonging to 10 different nationalities, resulting in highly diverse kitchen habits and cooking styles. Our dataset features 55 hours of video consisting of 11.5M frames, which we densely labelled for a total of 39.6K action segments and 454.2K object bounding boxes. Our annotation is unique in that we had the participants narrate their own videos (after recording), thus reflecting true intention, and we crowd-sourced ground-truths based on these.
We describe our object, action and
anticipation challenges, and evaluate several baselines over two test splits,
\textit{seen} and \textit{unseen} kitchens.
\added{We introduce new baselines that highlight the multimodal nature of the dataset and the importance of explicit temporal modelling to discriminate fine-grained actions (e.g. `closing a tap' from `opening' it up).}
\end{abstract}

}

\maketitle

\IEEEdisplaynontitleabstractindextext

% \fi
\IEEEpeerreviewmaketitle

\IEEEraisesectionheading{\section{Introduction}\label{sec:introduction}}

\begin{figure*}[t]
\centering
\includegraphics[width=1\linewidth]{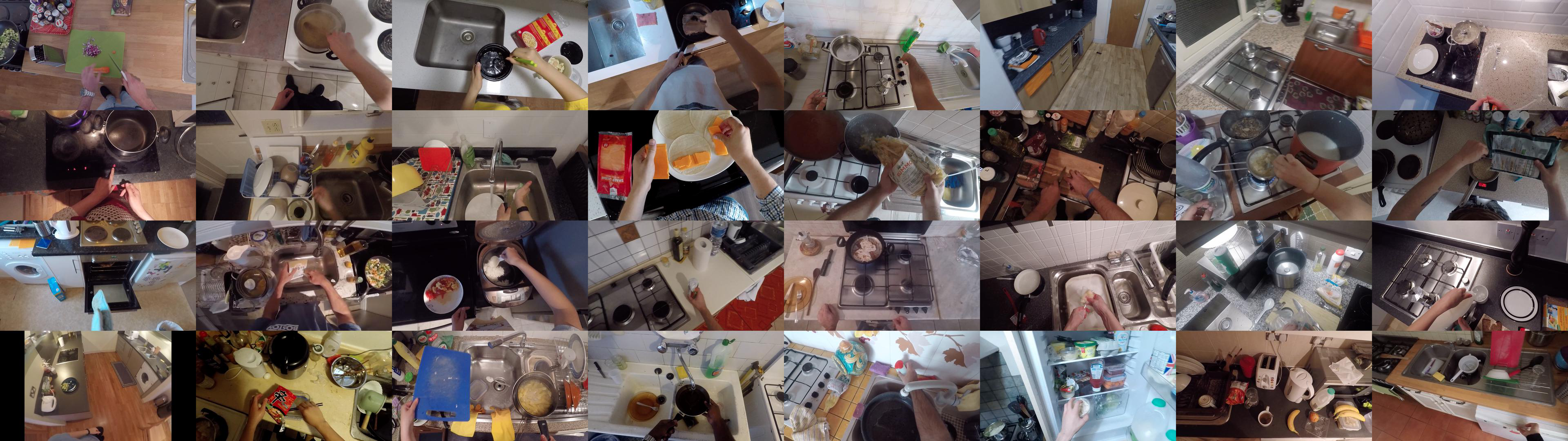}\\
\includegraphics[width=1\linewidth]{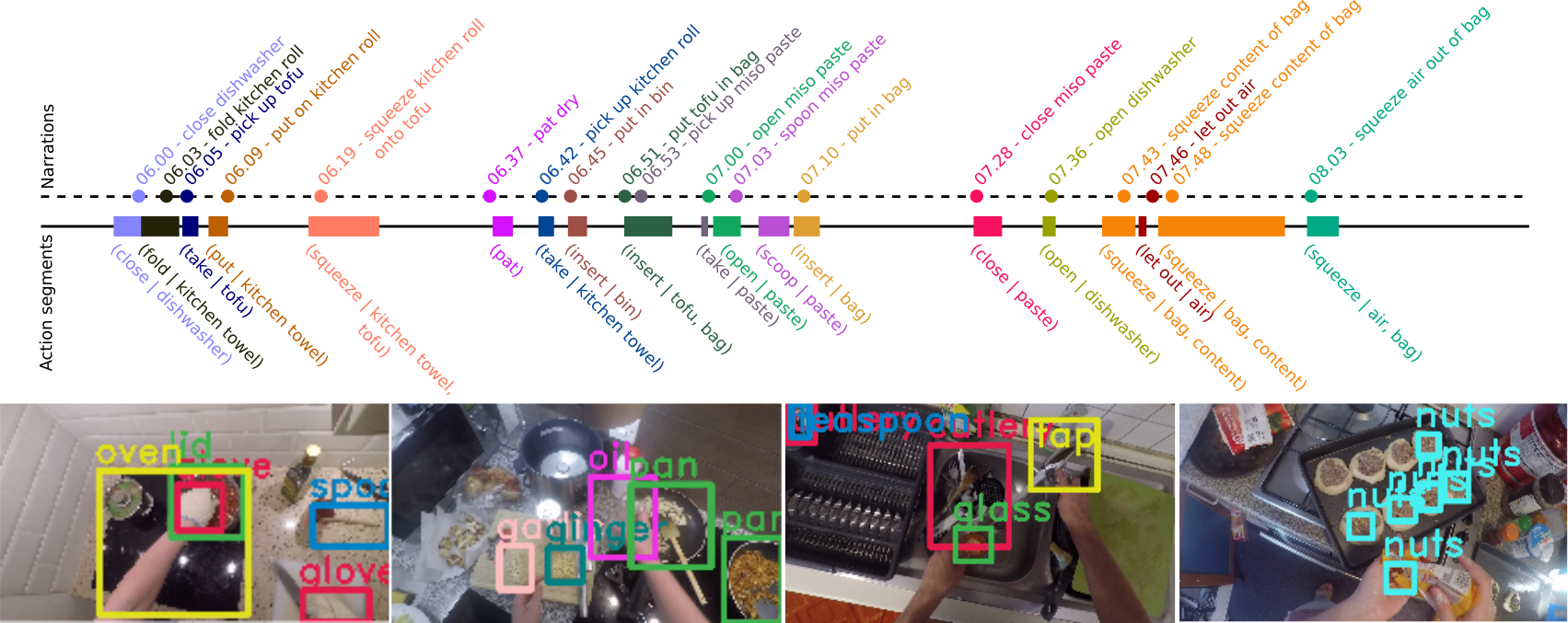}
\caption{\textbf{From Top}: Frames from the 32 environments; Narrations by participants used to annotate action segments; Active object bounding box annotations}
\label{fig:epic-wall}
\end{figure*}

In recent years, we have seen significant progress in many domains such as image classification~\cite{resnet}, object detection~\cite{ren2015faster}, captioning~\cite{Karpathy14} and visual question-answering~\cite{VQA}. This success has in large part been due to advances in deep learning~\cite{alexnet} as well as the availability of large-scale image benchmarks such as Pascal VOC~\cite{pascal}, ImageNet~\cite{imagenet}, MS-COCO~\cite{coco} and ADE~\cite{ade}.

While gaining attention, work in video understanding has been more scarce, mainly due to the lack of annotated datasets. This has been changing recently, with the release of the action classification benchmarks such as~\cite{Goyal2017,Haija2016,zhao2017slac,moviedescription,movieqa,fouhey2017lifestyle,Monfort2019moment,zhao2019hacs,carreira2019short}.
In~\cite{moviedescription}, the authors collected clips from movies for the task of video-based captioning, while~\cite{movieqa} evaluates story-based question-answering from videos.
With the exception of~\cite{movieqa}, most of these datasets contain videos that are very short in duration, i.e.~only a few seconds long, focusing on a single action. Hollywood in Homes~\cite{Sigurdsson2016} makes a step towards activity recognition by collecting 10K videos of humans performing various tasks in their home. While this dataset is a nice attempt to collect daily actions, the videos have been recorded in a scripted way, by asking AMT workers to act out a script in front of the camera. This makes the videos look oftentimes less natural, and they also lack the progression and multi-tasking of actions that occur in real life.

Here we focus on first-person vision, which offers a unique viewpoint on people's daily activities. This data is rich as it reflects our goals and motivation, ability to multi-task, and the many different ways to perform a variety of important, but mundane, everyday tasks (such as cleaning the dishes).
However, datasets to evaluate first-person vision algorithms~\cite{EGTEA,sigurdsson2018charadesego,Fathi2012,Damen2014a,de2008guide,Pirsiavash2012} have been significantly smaller in size than their third-person counterparts, often captured in a single environment~\cite{EGTEA,Fathi2012,Damen2014a,de2008guide}.
Daily interactions from wearable cameras are scarcely available online, making this a largely unavailable source. 

In this paper, we introduce \EPIC{}, a  large-scale egocentric dataset. Our data was collected by 32 participants, belonging to 10 nationalities, in their native kitchens (Fig.~\ref{fig:epic-wall}). The participants were asked to capture all their daily kitchen activities, and record sequences regardless of their duration. The recordings, which include both video and sound, not only feature the typical interactions with one's own kitchenware and appliances, but importantly show the natural multi-tasking that one performs, like washing a few dishes amidst cooking. Such parallel-goal interactions have not been captured in existing datasets, making this both a more realistic as well as a more challenging set of recordings. A video introduction to the recordings is available at: \textcolor{blue}{\underline{\url{http://youtu.be/Dj6Y3H0ubDw}}}.

Altogether, \EPIC{} has 55hrs of recording, densely annotated with start/end times for each action/interaction, as well as bounding boxes around objects subject to interaction. We describe our object, action and anticipation challenges, and report baselines in two scenarios, i.e., \textit{seen} and \textit{unseen} kitchens. We have released all of our data, and are tracking the community's progress on all challenges (with held out test ground-truth) via an online leaderboard. Details at: \textcolor{blue}{\underline{\url{http://epic-kitchens.github.io}}}.

\begin{table*}[t!]
\begin{center}
\caption{Comparative overview of relevant datasets. \small{$^*$action classes with $>50$ samples}}
\label{tab:datasets}
\resizebox{\linewidth}{!}{%
\begin{tabular}{|l|c|c|c|c|r|c|r|c|r|c|c|c|c|}
\hline

 & &\textbf{Non-} &\textbf{Native} & &  & \textbf{Sequ-} &\textbf{Action} & \textbf{Action} & \textbf{Object} & \textbf{Object}
& \textbf{Partici-} & \textbf{No.}\\
\textbf{Dataset} & \textbf{Ego?} &\textbf{Scripted?} &\textbf{Env?} &\textbf{Year} & \textbf{Frames} & \textbf{ences} & \textbf{Segments} &\textbf{Classes} & \textbf{BBs} & \textbf{Classes}
& \textbf{pants} &\textbf{Env.s} \\
\hline
\hline
\EPIC{} &$\checkmark$ &$\checkmark$ &$\checkmark$ &2018 &11.5M  &432  &39,596 &149* &454,158 &323  &32 & 32 \\
\hline
\hline
EGTEA Gaze+~\cite{EGTEA}  &$\checkmark$ &$\times$ &$\times$ &2018 & 2.4M
&86  &10,325  &106 &0 &0 &32 & 1 \\
Charades-ego~\cite{sigurdsson2018charadesego} &{\tiny 70\%} $\checkmark$ &$\times$ &$\checkmark$ &2018 &2.3M &2,751 & 30,516 &157 &0 &38 &71 &N/A \\
BEOID~\cite{Damen2014a} &$\checkmark$ &$\times$ &$\times$ &2014 &0.1M

&58  &742  &34 &0 &0 &5 & 1 \\
GTEA Gaze+~\cite{Fathi2012}  &$\checkmark$ &$\times$ &$\times$ &2012 & 0.4M

&35  &3,371  &42 &0 &0 &13 & 1 \\
ADL~\cite{Pirsiavash2012} &$\checkmark$ &$\times$ &$\checkmark$ &2012 &1.0M

&20  &436  &32 &137,780 &42 &20 &20 \\
CMU~\cite{de2008guide} &$\checkmark$ &$\times$ &$\times$ &2009 & 0.2M

&16  &516 &31 &0 &0 &16 &1 \\
\hline\hline
VLOG \cite{fouhey2017lifestyle} & $\times$ & $\checkmark$ & $\checkmark$ & 2017 & 37.2M & 114K & 0 & 0 & 0 & 0 & 10.7K & N/A\\
Charades \cite{Sigurdsson2016} &$\times$ &$\times$ &$\checkmark$ &2016 &7.4M & 9,848 &67,000  & 157 &0 &0 & N/A & 267\\
Breakfast \cite{Kuehne2014}  &$\times$ &$\checkmark$ &$\checkmark$ &2014 &
3.0M

& 433 & 3078 & 50 &0 &0 & 52 & 18 \\
50 Salads \cite{Stein2013} &$\times$ &$\times$ &$\times$ &2013 & 0.6M

& 50 & 2967 & 52 &0 &0 & 25 & 1 \\
MPII Cooking 2 \cite{Rohrbach2012} &$\times$ &$\times$ &$\times$ &2012 & 2.9M

& 273 & 14,105 & 88 &0 &0 & 30 & 1 \\
\hline
\end{tabular}}
\end{center}

\end{table*}

\section{Related Datasets}
We compare \EPIC{} to six commonly-used egocentric datasets~\cite{EGTEA,Fathi2012,sigurdsson2018charadesego,Damen2014a,de2008guide,Pirsiavash2012} in Table~\ref{tab:datasets}, as well as five
third-person activity-recognition datasets \cite{fouhey2017lifestyle,Sigurdsson2016,Kuehne2014,Stein2013,Rohrbach2012} that focus on object-interaction activities.
We exclude  egocentric datasets that focus on inter-person interactions~\cite{Alletto2015,Fathi2012b,Ryoo2013}
as well as instructional videos \cite{Zhou2017,alayrac16objectstates,Doughty2019Pros,Miech2019how}, as these target different research questions.

A few datasets aim at capturing activities in native environments, most of which are recorded in third-person~\cite{Goyal2017,fouhey2017lifestyle,Sigurdsson2016,sigurdsson2018charadesego,Kuehne2014}.~\cite{Kuehne2014} focuses on cooking dishes based on a list of breakfast recipes. In~\cite{fouhey2017lifestyle}, videos depicting interactions with 30 daily objects are collected by querying YouTube, while~\cite{Goyal2017,Sigurdsson2016} are scripted -- subjects are requested to enact a crowd-sourced storyline~\cite{Sigurdsson2016} or a given action~\cite{Goyal2017}, which oftentimes results in less natural looking actions.
Most egocentric datasets similarly use scripted activities, i.e. people are told what actions to perform. When following  instructions, participants perform steps in a sequential order, as opposed to the more natural real-life scenarios addressed in our work, which involve multi-tasking, searching for an item, thinking what to do next, changing one's mind or even unexpected surprises.

\EPIC{} is most closely related to the ADL dataset~\cite{Pirsiavash2012} which also provides egocentric recordings in native environments. However, our dataset is substantially larger: it has 11.5M frames vs 1M in ADL, 90x more annotated action segments, and 4x more object bounding boxes, making it the largest first-person dataset to date. 

\section{The \EPIC{} Dataset}
\label{sec:dataset}

In this section, we describe the collection and annotation pipeline, and present statistics, showcasing different aspects of the dataset.

\subsection{Data Collection}
\label{sec:recording}

\begin{figure}[t]
{\includegraphics[width=\columnwidth]{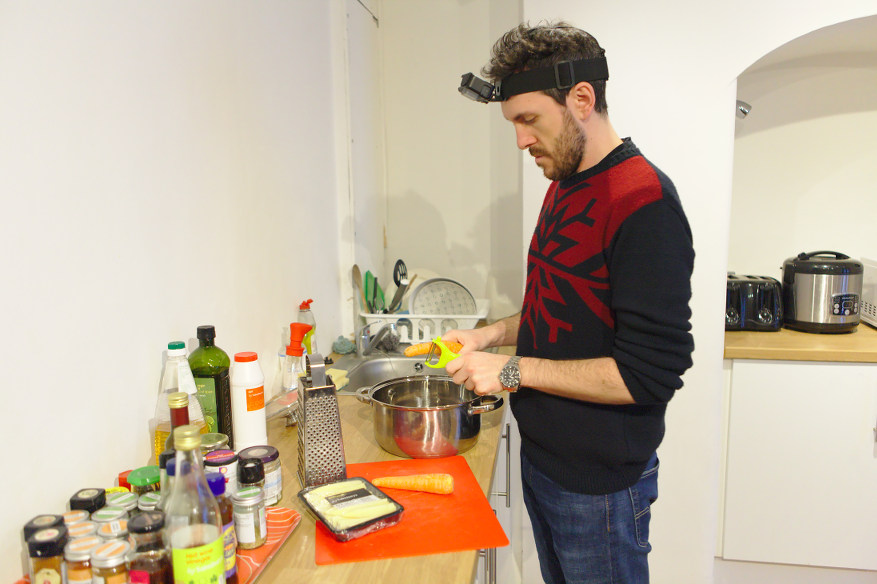}}
\caption{\added{Head-mounted GoPro used in dataset recording}}
\label{fig:setup}
\end{figure}

The dataset was recorded by 32 individuals in 4 cities in different countries (in North America and Europe): 15 in Bristol/UK,
8 in Toronto/Canada,
8 in Catania/Italy
and 1 in Seattle/USA
between May and Nov~2017.
Participants were asked to capture all kitchen visits for at least \textit{three consecutive days}, with the recording starting immediately before entering the kitchen,
allowing a few seconds to ensure the camera starts before carrying out the daily activities,
and only stopped before leaving the kitchen.
They recorded the dataset voluntarily and were not financially rewarded. The participants were asked to
be in the kitchen alone for all the recordings, thus capturing only one-person activities. We also asked them to remove all items that would disclose their identity such as portraits or mirrors.

Data was captured using a head-mounted Go-Pro with an adjustable mounting to control the viewpoint for different environments and participants' heights.
\added{Fig.~\ref{fig:setup} demonstrates the mounting which ensures desktop interactions are within the capturing field of view.}
Before each recording, the participants checked the battery life and viewpoint, using the GoPro Capture mobile app,
so that their outstretched hands were approximately located at the middle of the camera frame. 
 \revision{Stereo audio was captured from the GoPro's built-in microphone, at a sampling rate of 48000kHz and a bit rate of 128kb/s.}
The camera was set to linear field of view~(fov), 59.94\fps{} and Full HD resolution of 1920x1080, however some subjects made minor changes like wide or ultra-wide fov or resolution, as they recorded multiple sequences in their homes, and thus were switching the device off and on over several days.
Specifically, 1\% of the videos were recorded at 1280x720
and 0.5\% at 1920x1440. Also, 1\% at 30\fps, 1\% at 48\fps~ and 0.2\% at 90\fps\footnote{\revision{The videos are publicly available including the 2-channel audio stream. We also provide preprocessed frames suitable for direct use in standard video architectures.
  We offer RGB JPEG frames at a resolution of 456x256 resampled to 60\fps{}.
  From these frames we also compute $\tvl$ optical flow~\cite{zach2007duality} at 30\fps{} quantized into grayscale JPEGs.}}.

\begin{figure}[t!]
\begin{tcolorbox}[width=\linewidth,boxrule=1pt,colback=blue!4,left=2pt,right=2pt,top=2pt,bottom=2pt]
\scriptsize{
\noindent Use any word you prefer. Feel free to vary your words or stick to a few.\\[-2.3mm]

\noindent Use present tense verbs (e.g. cut/open/close).\\[-2.3mm]

\noindent Use verb-object pairs (e.g. “wash carrot”).\\[-2.3mm]

\noindent You may (if you prefer) skip articles and pronouns (e.g. ``cut kiwi'' rather than ``I cut the kiwi'').\\[-2.3mm] 

\noindent Use propositions when needed (e.g. ``pour water into kettle'').\\[-2.3mm]

\noindent Use `and' when actions are co-occurring (e.g. ``hold mug and pour water'').\\[-2.3mm]

\noindent If an action is taking a long time, you can narrate again (e.g. ``still stirring soup'').}
\end{tcolorbox}
\caption{Instructions used to collect video narrations from our participants}
\label{fig:narrationInstruction}
\end{figure}

\begin{figure*}
\includegraphics[width=0.25\textwidth]{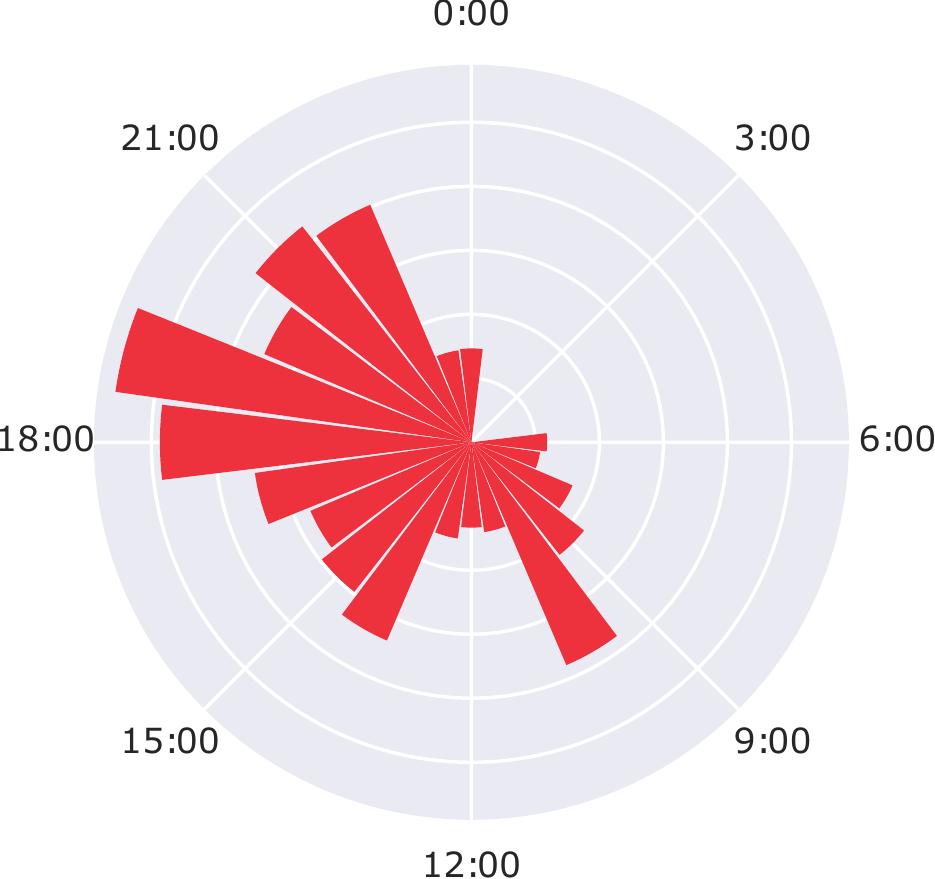}
\includegraphics[width=0.45\textwidth]{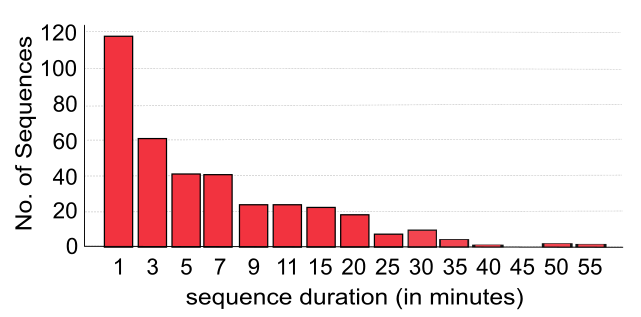}
\includegraphics[width=0.28\textwidth]{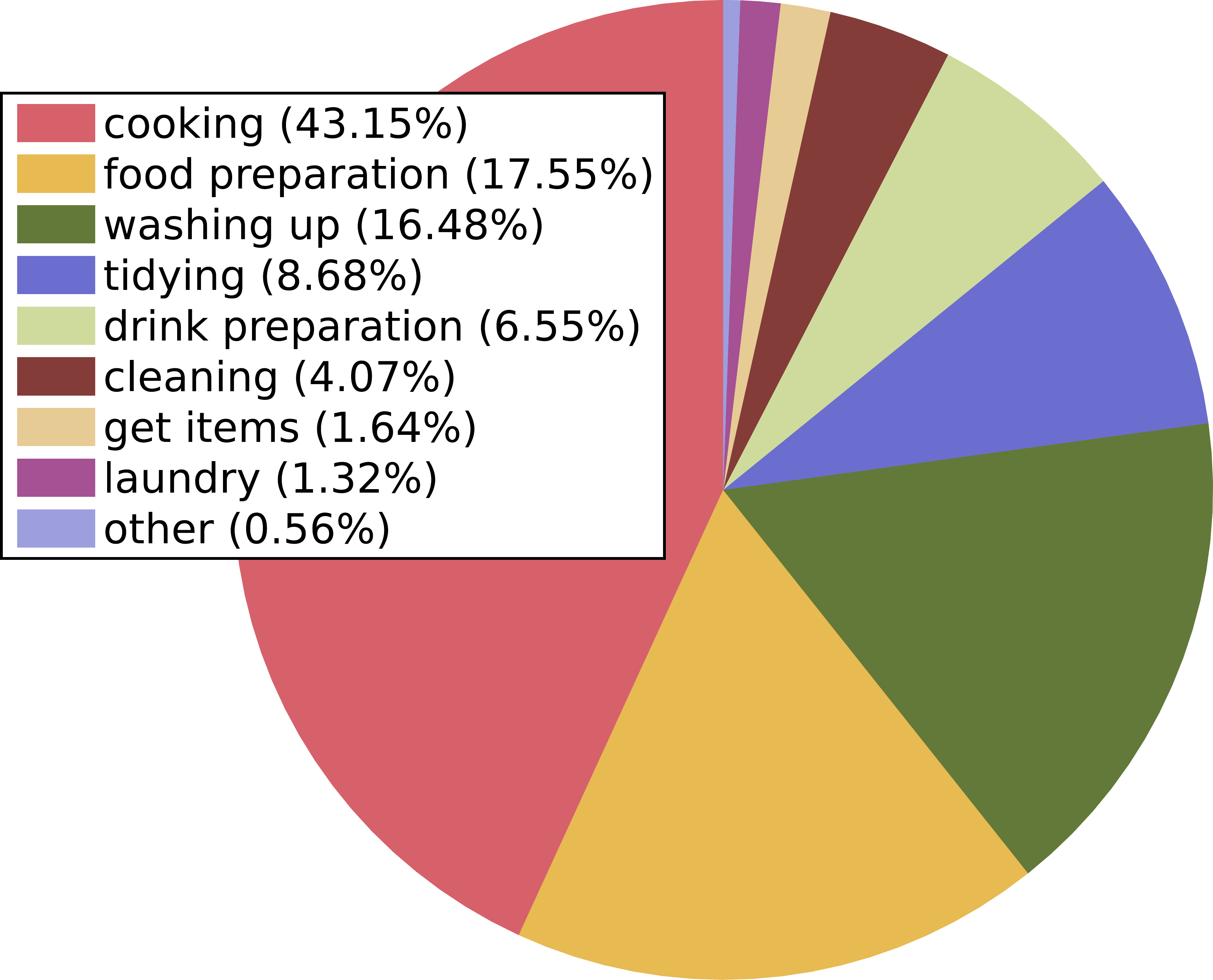}\\
{\includegraphics[width=0.2\textwidth]{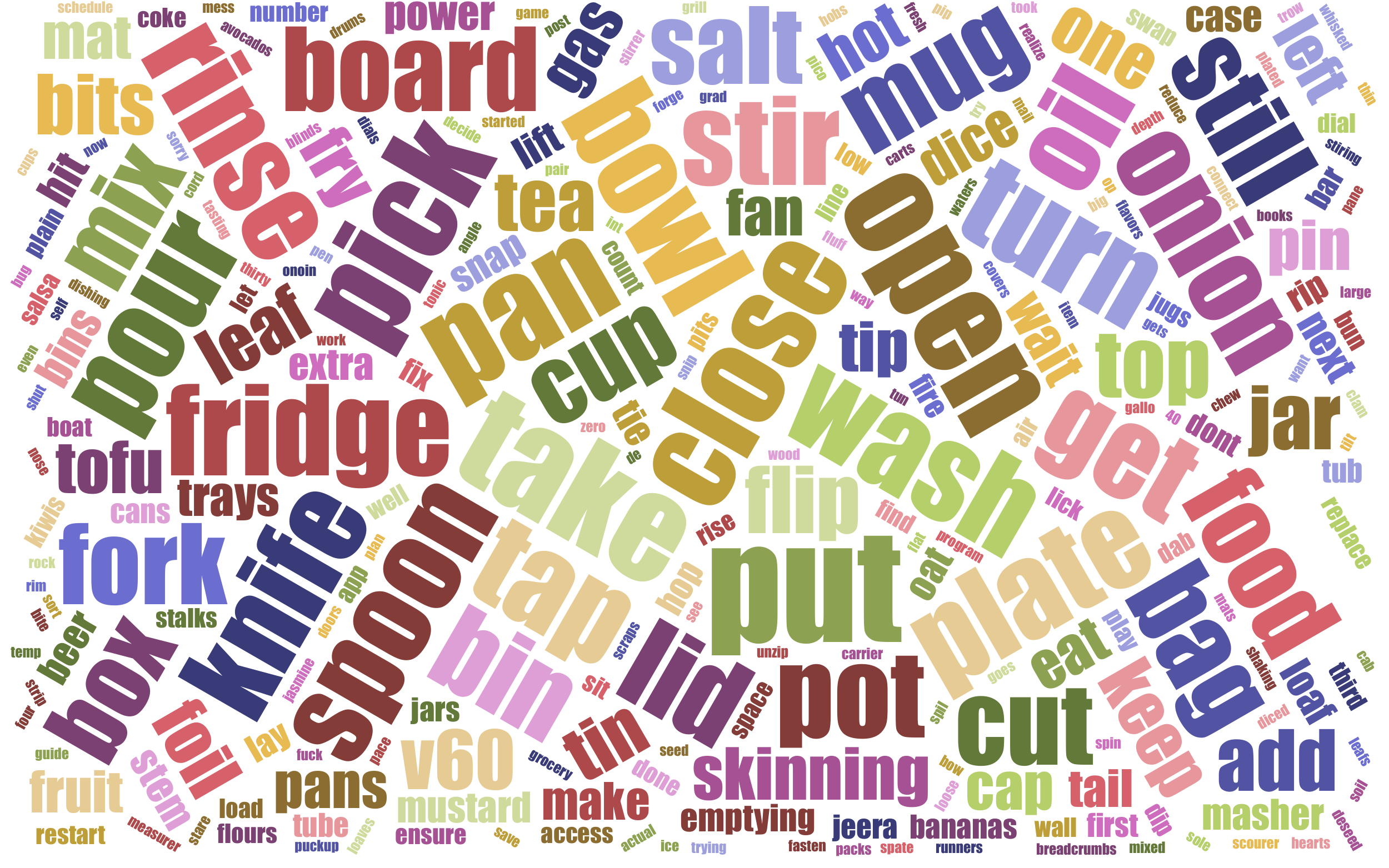}}
\subfloat 
{\includegraphics[width=0.2\textwidth]{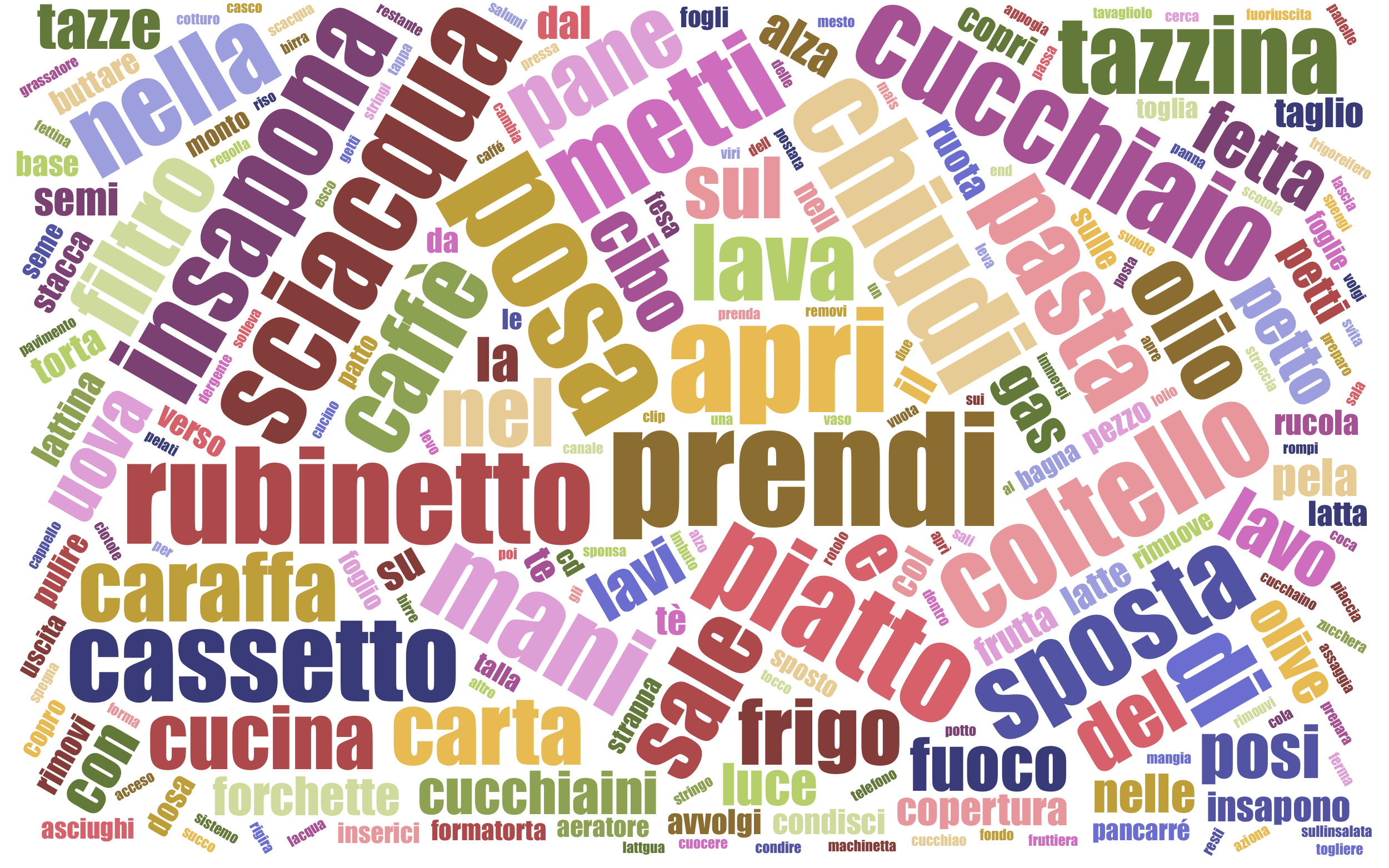}}
\subfloat 
{\includegraphics[width=0.2\textwidth]{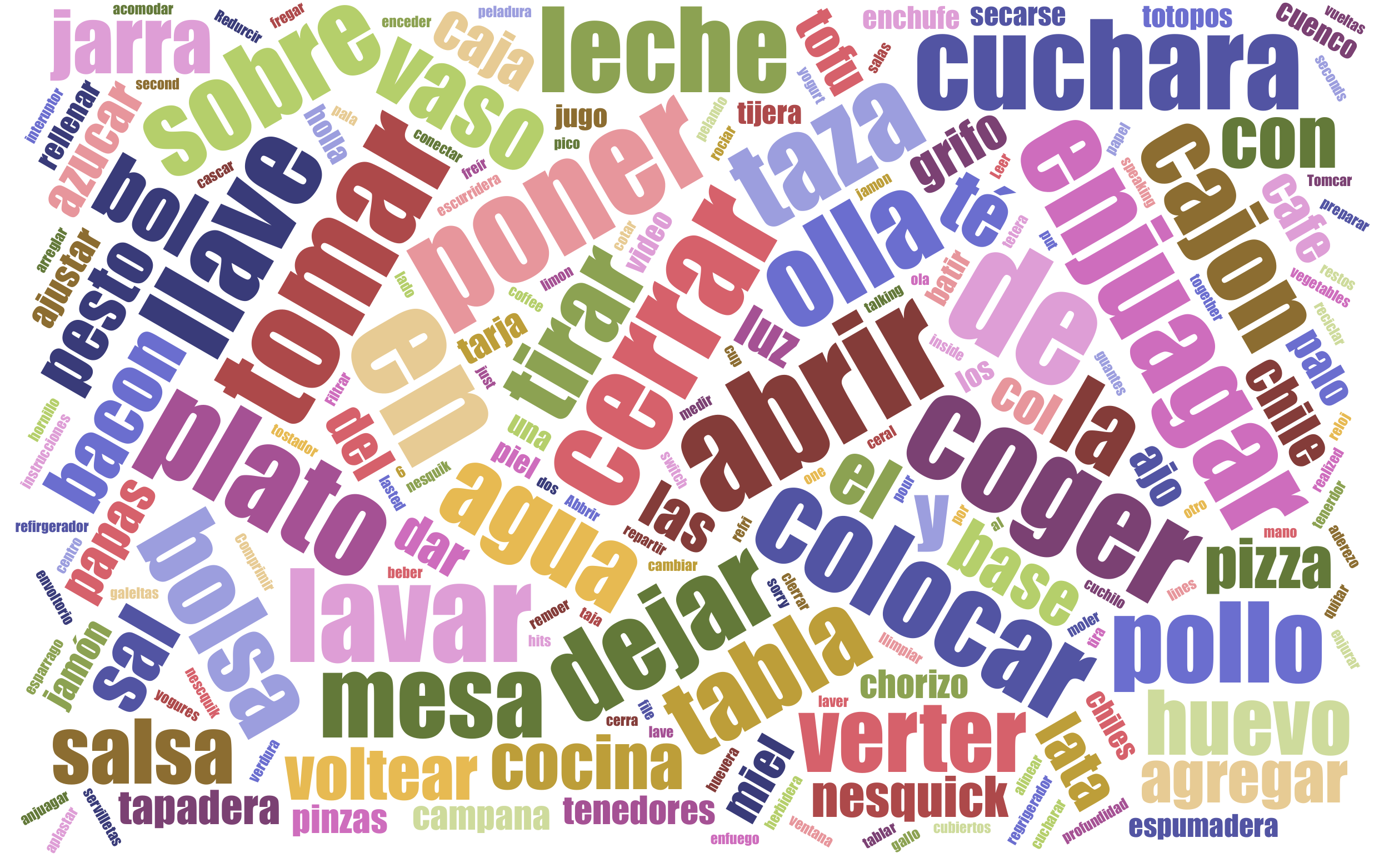}}
\subfloat
{\includegraphics[width=0.2\textwidth]{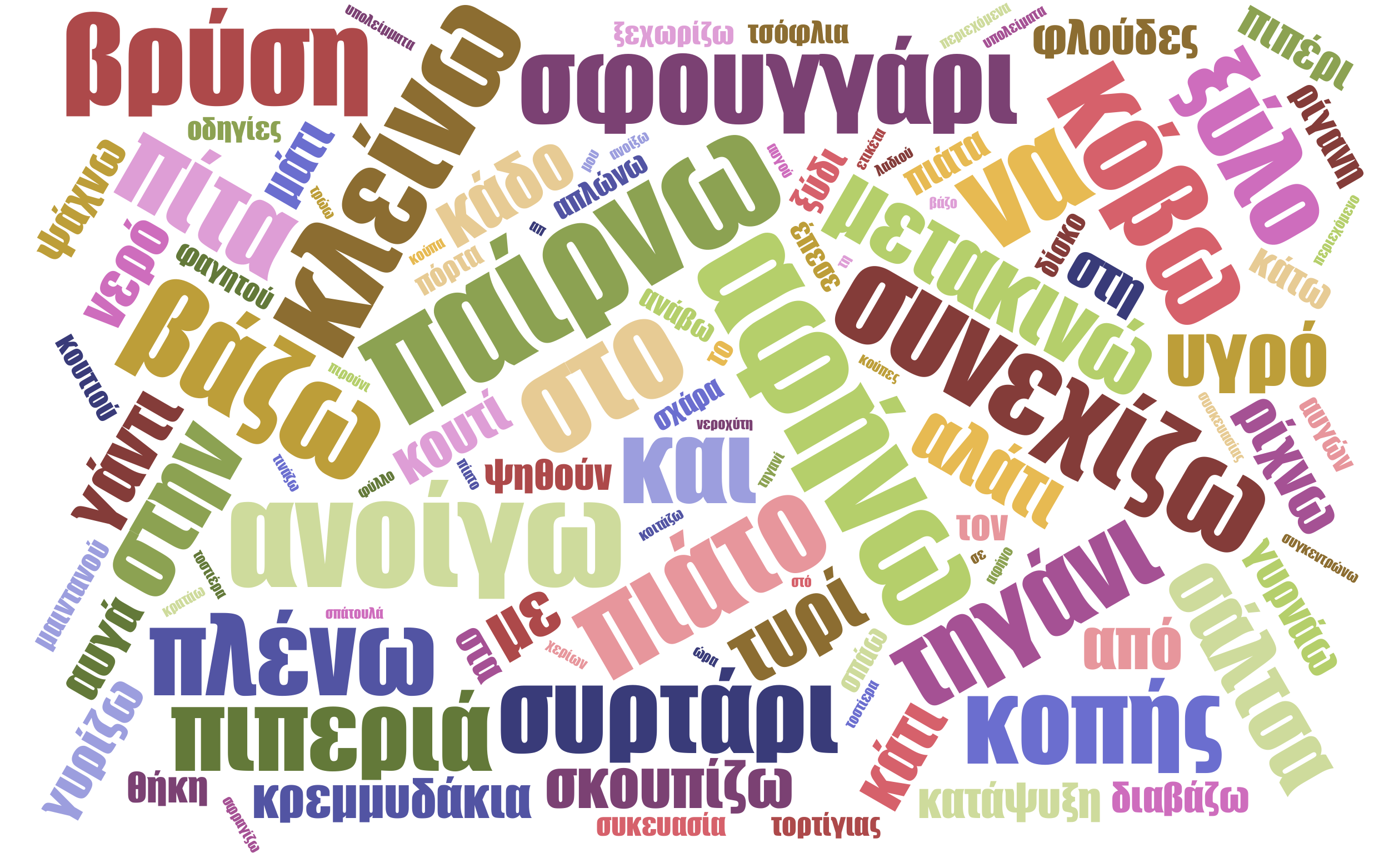}}
\subfloat
{\includegraphics[width=0.2\textwidth]{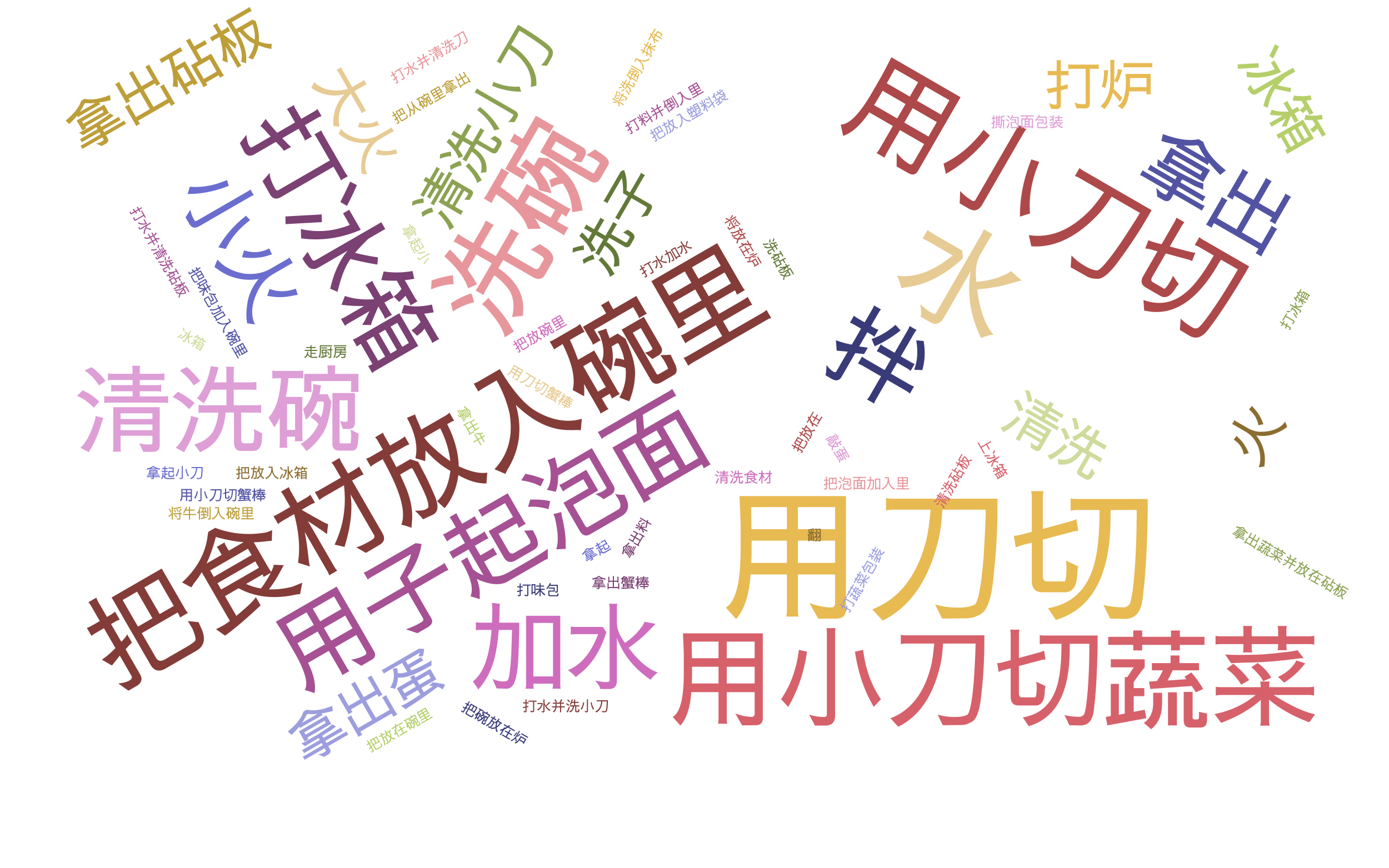}}
\caption{{\bf Top} (left to right): time of day of the recording, histogram of sequence durations and pie chart of high-level goals; {\bf Bottom}: Wordles of narrations in native languages (English, Italian, Spanish, Greek and Chinese)}
\label{fig:statsCollect}
\end{figure*}

The recording lengths varied depending on the participant's kitchen engagement.
On average, people recorded for 1.7hrs, with the maximum being 4.6hrs and the minimum just over half an hour.
Cooking a single meal can span multiple sequences, depending on whether one stays in the kitchen, or leaves and returns later. On average, each participant recorded 13.6 sequences.
Figure~\ref{fig:statsCollect} presents statistics on time of day using the local-time of the recording, high-level goals and sequence durations.

Since crowd-sourcing annotations for such long videos is very challenging, we had our participants do a coarse first annotation themselves.
Each participant was asked to watch their videos, after completing all recordings, and narrate the actions carried out, using a hand-held recording device. We opted for a \revision{speech} recording rather than written captions as this is much faster for the participants, who were thus more willing to provide these annotations\footnote{A freely-available application on smart phone was used to gather the narrations' recordings.}.  \revision{Throughout this paper, we refer to the post-filming narration recording as ``speech'' to distinguish it from the ``audio'' stream captured naturally from the sounds of actions recorded by the GoPro camera.}
Recent attempts in image annotations using speech report a speed-up of up to 15x in annotating ImageNet when spoken annotations were acquired~\cite{Gygli2019efficient}.
These are analogous to a \textit{live commentary} of the video. The general instructions for narrations are listed in Fig.~\ref{fig:narrationInstruction}.
The participant narrated in English if sufficiently fluent or in their native language. In total, 5 languages were used: 17 narrated in English, 7 in Italian, 6 in Spanish, 1 in Greek and 1 in Chinese. Figure~\ref{fig:statsCollect} shows wordles of the most frequent words in each language.

\begin{table*}[t]
\caption{Extracts from 6 transcription files in .sbv format}
\resizebox{\textwidth}{!}{
\begin{tabular}{ |l|l|l|l|l|l| }
\hline
 0:14:44.190,0:14:45.310 & 0:00:02.780,0:00:04.640 & 0:04:37.880,0:04:39.620 & 0:06:40.669,0:06:41.669 & 0:12:28.000,0:12:28.000 & 0:00:03.280,0:00:06.000 \\
 pour tofu onto pan & open the bin & take onion & pick up spatula & pour pasta into container & open fridge\\
 0:14:45.310,0:14:49.540 & 0:00:04.640,0:00:06.100 & 0:04:39.620,0:04:48.160 & 0:06:41.669,0:06:45.250 & 0:12:33.000,0:12:33.000 & 0:00:06.000,0:00:09.349 \\
 put down tofu container & pick up the bag & cut onion & stir potatoes & take jar of pesto & take milk \\
 0:14:49.540,0:15:02.690 & 0:00:06.100,0:00:09.530 & 0:04:48.160,0:04:49.160 & 0:06:45.250,0:06:46.250 &0 :12:39.000,0:12:39.000 & 0:00:09.349,0:00:10.910 \\
 stir vegetables and tofu & tie the bag & peel onion & put down spatula & take teaspoon & put milk \\
 0:15:02.690,0:15:06.260 & 0:00:09.530,0:00:10.610 & 0:04:49.160,0:04:51.290 & 0:06:46.250,0:06:50.830& 0:12:41.000,0:12:41.000 &0:00:10.910,0:00:12.690  \\
 put down spatula & tie the bag again & put peel in bin & turn down hob & pour pesto in container & open cupboard \\
 0:15:06.260,0:15:07.820 & 0:00:10.610,0:00:14.309 & 0:04:51.290,0:05:06.350 & 0:06:50.830,0:06:55.819 & 0:12:55.000,0:12:55.000 & 0:00:12.690,0:00:15.089 \\
 take tofu container & pick up bag &  peel onion & pick up pan & place pesto bottle on table &take bowl \\
 0:15:07.820,0:15:10.040 & 0:00:14.309,0:00:17.520 & 0:05:06.350,0:05:15.200 & 0:06:55.819,0:06:57.170 & 0:12:58.000,0:12:58.000 & 0:00:15.089,0:00:18.080 \\
 throw something into the bin & put bag down & put peel in bin & tip out paneer & take wooden spoon & open drawer \\
 \hline
\end{tabular}}
\label{fig:transcriptfiles}
\end{table*}

\begin{table*}[t]
\centering
\caption{\added{Sample Video Summaries}}
\resizebox{1\textwidth}{!}{
\begin{tabular}{ |l|l||l|l| }
\hline
P04-04.mp4 &making curries - fried paneer, boiled potatoes, chopped veg &
P07-08.mp4 &pour coffee and prepare tortilla with cheese and pepperoni\\
P13-08.mp4 &clean the dishes and prepare spaghetti carbonara &
P19-04.mp4 &made steamed noodles with beans, tomatoes, chicken\\
P23-02.mp4 &cooking Indian egg curry while cleaning dishes&
P28-09.mp4 &prepare avocado and tomato salad\\
\hline
\end{tabular}}
\label{tab:video-summaries}
\end{table*}

Our decision to collect narrations from the participants themselves is because they are the most qualified to label the activity compared to an independent observer, as they were the ones performing the actions. We opted for a post-recording narration such that the participant performs her/his daily activities undisturbed, without being concerned about labeling.

We tested several automatic \revision{speech}-to-text APIs~\cite{googleSpeechApi,ibmWatson,cmuSphinx},
which failed to produce accurate transcriptions as these expect a relevant corpus and complete sentences for context.
We thus collected manual transcriptions via Amazon Mechanical Turk~(AMT), and used YouTube's automatic closed caption alignment tool to produce accurate timings. For non-English narrations, we also asked AMT workers to translate the sentences.
To make the job more suitable for AMT, \revision{speech} files are split by removing silence below a pre-specified decibel threshold (after compression and normalisation).  Speech chunks are then \revision{grouped into AMT's Human Intelligent Tasks (HITs)}\footnote{HITs are groupings of tasks within AMT, where the crowdsourced turker is required to complete all tasks within the HIT before they submit for payment.} with a duration of around 30 seconds each. 
To ensure consistency, we submit the same HIT three times and select the ones with an edit distance of 0 to at least one other HIT. We manually corrected cases when there was no agreement. 
Examples of transcribed and timed narrations are provided in Table~\ref{fig:transcriptfiles}.
\added{The participants were also asked to provide one sentence per sequence describing the overall goal or activity that took place (examples in Table~\ref{tab:video-summaries}).}

In total,
reporting translated or originally-English narrations,
we collected $39,596$ action narrations, corresponding to a narration every $4.9s$ in the video. The average number of words per phrase is $2.8$ words.
These narrations give us an initial labeling of all actions with rough temporal alignment (obtained from the timestamp of the \revision{speech} with respect to the video). However, narrations are also not a perfect source of ground-truth:

\begin{itemize}[leftmargin=*]
\item The narrations can be incomplete, i.e., the participants were selective in which actions they chose to narrate.
We noticed that they labeled the `open' actions more than their counter-action `close', as the narrator's attention has already moved to the next goal. We consider this phenomena in our evaluation, by only evaluating actions that have been narrated.

\item Temporally,
the narrations are belated, after the action takes place. This is adjusted using ground-truth action segments (see Sec.~\ref{subsec:strongAction}).
\item Participants use their own vocabulary and free language. While this is a challenging issue to deal with in evaluation, we believe it is important to push the community to go beyond the pre-selected list of labels in the future (also argued in~\cite{ade}). 
We here resolve this issue by grouping verbs and nouns into  
minimally overlapping classes (see Sec.~\ref{subsec:clusters}). 
\end{itemize}

\begin{figure*}[t]
\includegraphics[width=\linewidth]{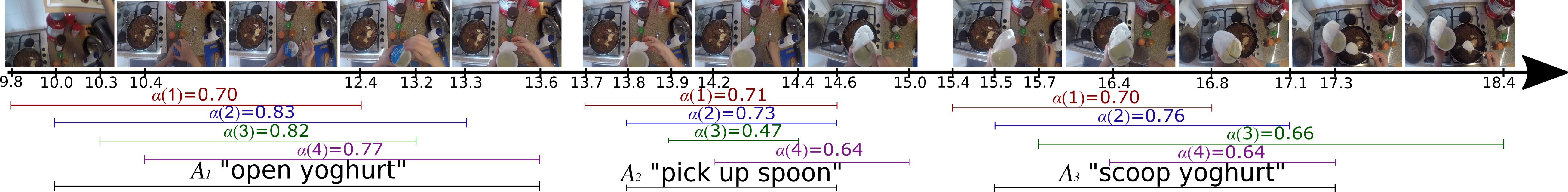}
\caption{Example temporal annotations for 3 consecutive actions.}
\label{fig:annotations_example}
\end{figure*}

\subsection{Action Segment Annotations}
\label{subsec:strongAction}

For each narrated sentence, we adjust the start and end times of the action using AMT.
To ensure the annotators are trained to perform temporal localisation, we include an introductory tutorial based on our previous work's understanding~\cite{Moltisanti2017} that explains temporal bounds of actions.
Each HIT is composed of a maximum of 10 consecutive narrated phrases $p_i$, where annotators label $A_{i} = [t_{s_i}, t_{e_i}]$ as the start and end times of the $i^{th}$ action.
Two constraints were added to decrease the amount of noisy annotations: First, the action has to be at least 0.5 seconds in length;
second, the action cannot start before the preceding action's start time.
Note that consecutive actions are allowed to overlap.
Moreover, the annotators could indicate that the action does not appear in the video. This handles occluded, impossible to distinguish or out-of-bounds cases.

To ensure consistency, we ask $\mathcal{K}_a=4$ annotators to annotate each HIT. We filter, reject and resubmit unacceptable hits through a combination of automatic and manual checks.
\added{
From $A_i(j)$ ($i$~indexes the action and $j$ indexes the annotator),
we calculate the ground-truth action segments $A_i$, where we considered 6 different consensus functions:}

\noindent
\added{
a) \textit{average start/end}:
\begin{equation}
    A_i=[\text{mean}(t_{s_i}(j)), \text{mean}(t_{e_i}(j))],
\end{equation}
}
\added{
\noindent b) \textit{median start/end}:
\begin{equation}
    A_i=[\text{median}(t_{s_i}(j)), \text{median}(t_{e_i}(j))],
\end{equation}
}
\added{
\noindent c) \textit{minimum duration}, where we select the annotator that provided the shortest annotation as:
\begin{equation}\label{eq:min_dur}
    \jhat=\argmin_j t_{e_i}(j)-t_{s_i}(j),
\end{equation}
}
\added{
\noindent d) \textit{maximum duration}, where we select the annotator that provided the longest annotation as:
\begin{equation}\label{eq:max_dur}
    \jhat=\argmax_j t_{e_i}(j)-t_{s_i}(j),
\end{equation}
}
\added{
\noindent e) \textit{maximum agreement}.  Here, we first
calculate the inter-annotator agreement for each annotation $A_i(j)$:
\begin{equation}
\alpha_i(j) = \frac{1}{K_a} \sum_{k=1}^{\mathcal{K}_a} \mathrm{IoU} (A_i(j), A_i(k))
\end{equation}
and then find the annotator with the maximum agreement:
\begin{equation}\label{eq:max_aggr}
    {\jhat = \argmax_j \alpha_i(j)}\text{.}
\end{equation}
}
\added{
In c), d), e), the ground-truth annotation is finally selected to be $A_i=A_i(\jhat)$. Finally, we consider:
}

\added{
\noindent f) \textit{union of best agreements}, which is an extension of e). Additionally to \eqref{eq:max_aggr}, we find ${\hat{k} = \argmax_k \mathrm{IoU}(A_i(\jhat), A_i(k))}$, and then
}
\added{
the ground-truth $A_i$ is defined as:
\begin{equation}
\small{
A_i =\begin{cases}
\mathrm{union}(A_i(\jhat), A_i(\hat{k})), & \text{if  IoU} (A_i(\jhat), A_i(\hat{k}))>0.5\\
A_i(\jhat), & \text{otherwise.}
\end{cases}}
\label{eq:actionunion}
\end{equation}
a)-d) were strongly affected by outlier annotations, e.g. too short or too long action segments, and the decision was not made based on the agreement between annotators. e) and f) were the two best performing consensus functions. We finally chose f) for computing ground-truth temporal bounds of an action, where two annotations are combined when they have a strong agreement, since in some cases the single (best) annotation e) results in a too tight of a segment.
Numerically, the average action length increased from 3.5 sec using e) to 3.7 sec with f).
}
Figure~\ref{fig:annotations_example} shows examples of combining annotations.

\begin{figure}[t]
\centering
\includegraphics[width=1\linewidth]{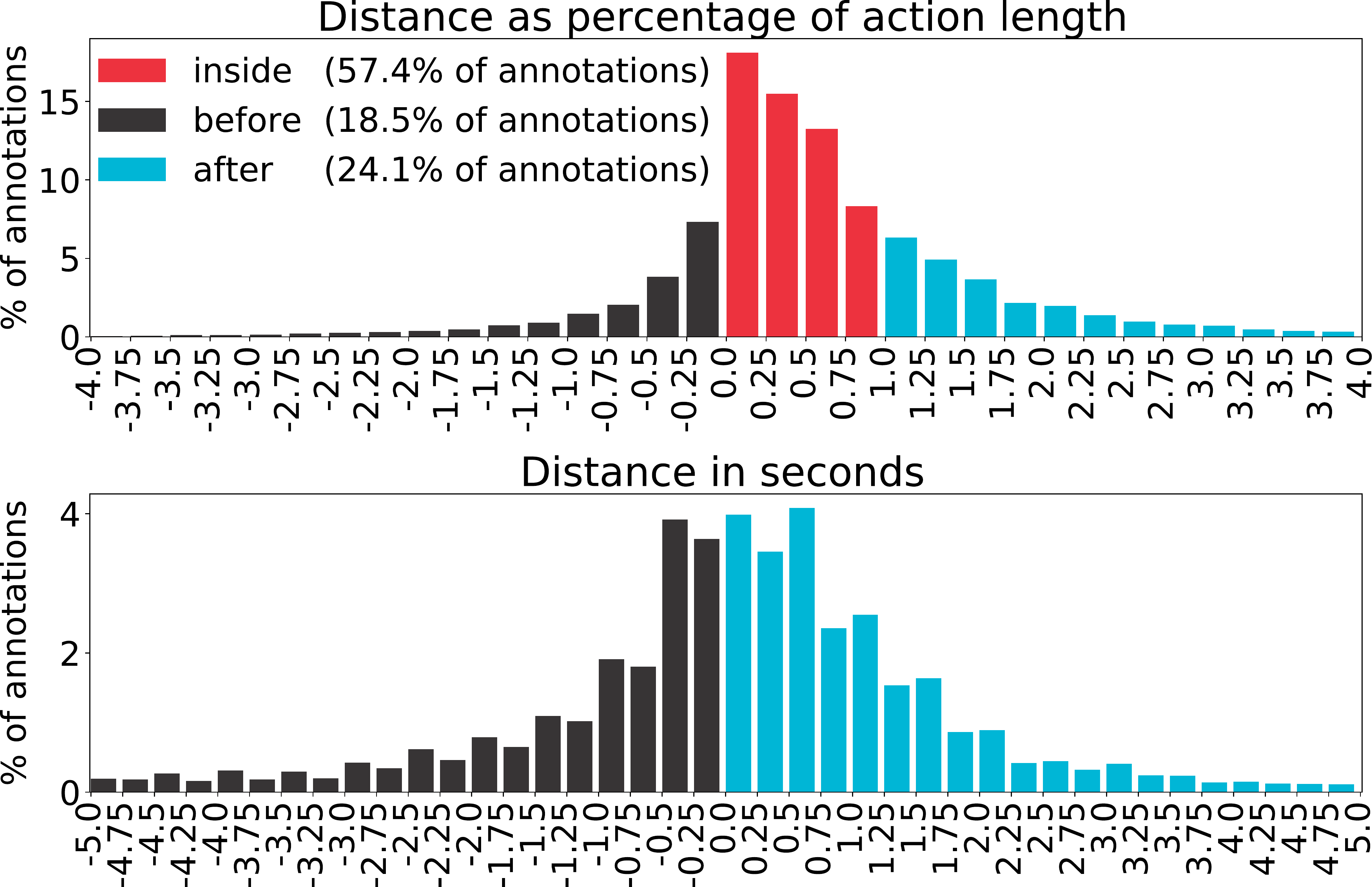}
\caption{\added{Distance between narration timestamp and ground truth action segments. Top: normalised using the action segment length.
Bottom: in seconds.
The narration timestamp is relative to the \revision{speech recorded} by the participants. Ground truth segments refer to the video start and end times.}}
\label{fig:narration_to_gt_distance}
\end{figure}

\added{Figure~\ref{fig:narration_to_gt_distance} illustrates the distance between the narration timestamp and the ground truth action segments.
The narration timestamp is relative to the \revision{speech recorded} by the participants. Ground truth segments refer to the video start and end times labelled from AMT annotations and the method in Eq~\ref{eq:actionunion}.
In the top plot the distance is normalised by the segment's length.
Red bars depict narrations whose timestamp is enclosed by the corresponding action segment (values between 0 and 1).
Black and blue bars indicate narrations that are respectively before and after the action's start and end, with negative values indicating that the narration timestamp is before the action.
In the bottom plot the distance is reported in seconds.
Plots show that most narrations are close to the action's segment (within 25\% of its length), with more narrations being belated rather than anticipated. This is natural given that narrators have to first recognise the action before narrating it, although in many cases the narrators were also able to predict their own actions.}

\begin{figure*}[t]
\includegraphics[width=\textwidth]{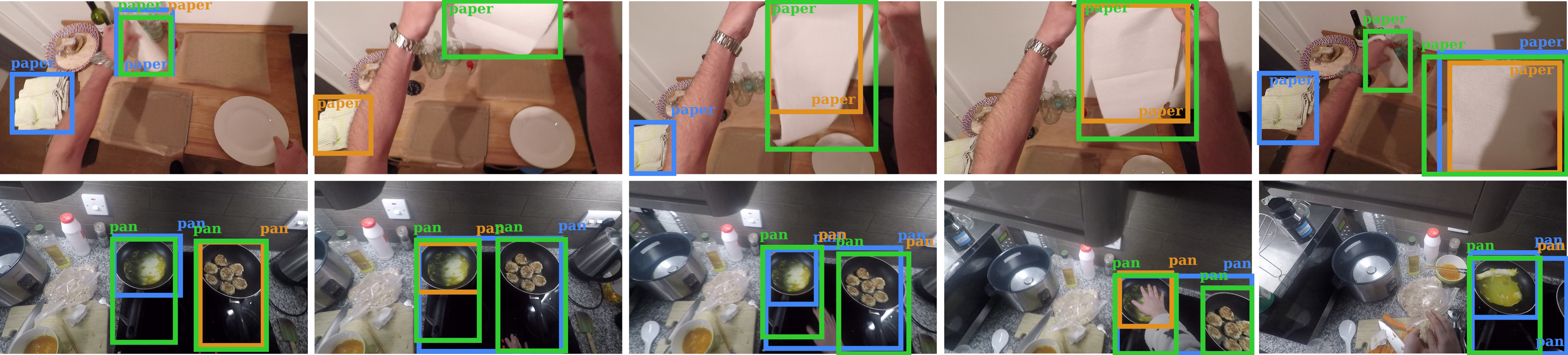}
\caption{Object annotation selection, our method (green) vs. best annotation per frame (orange). Other annotators are in blue.}
\label{fig:object_ann}
\end{figure*}

\added{In total, 18.5\% and 24.1\% of narrations were narrated before and after the action, while 57.4\% of the narrations were contained within their action segment. We also analyse the number of narrations that were contained in a different action segment. Overall, 30.6\% were contained within a different action. This shows the challenge involved in annotating actions starting from narration timestamps, as well as the challenges involved in using these timestamps as weak temporal supervision as recently showed in~\cite{moltisanti19action}.}

In total, we collected ground-truth start and end labels, for 39,564 action segments (lengths: $\mu=3.7s$, $\sigma=5.6s$). These represent 99.9\%  of all narrations. The missed annotations were those labelled as ``not visible'' by the annotators (but mentioned in narrations).
\added{Of these, 9,495 segments (24\%) overlap with one other action segment, highlighting concurrent interactions in natural recordings.}

\subsection{Active Object Bounding Box Annotations}
\label{subsec:strongObject}

The narrated \textit{nouns} correspond to objects relevant
to the action~\cite{Lee2012,Damen2014a}. Assume $\mathcal{O}_i$ is the set of one or more nouns in the phrase $p_i$ associated with the action segment $A_i = [t_{s_i}, t_{e_i}]$.
We consider each frame $f$ within $[t_{s_i}-2s,t_{e_i}+2s]$ as a potential frame to annotate the bounding box(es), for each object in $\mathcal{O}_i$.
We build on the interface from~\cite{tangseng2017} for annotating bounding boxes on AMT.
Each HIT aims to get an annotation for one object, for the maximum duration of $25s$, which corresponds to $50$ consecutive frames at $2$\fps.
The annotator can also note that the object does not exist in~$f$.
We particularly ask the same annotator to annotate consecutive frames to avoid subjective decisions on the extents of objects.
We also assess annotators' quality by ensuring that the annotators obtain an ${\textrm{IoU} \ge 0.7}$ on two golden annotations at the start of every HIT.
We request $\mathcal{K}_o = 3$ workers per HIT, and select the one with maximum agreement $\beta$:
\begin{equation}
\beta(q) = \sum_f \max\limits_{j \ne q}^{\mathcal{K}_o}\, \max_{k,l}\, \text{IoU}(\text{BB}(j, f, k), \text{BB}(q, f, l))
\label{eq:overallBestAnnotator}
\end{equation}

\begin{figure*}[t]
\includegraphics[width=1\textwidth]{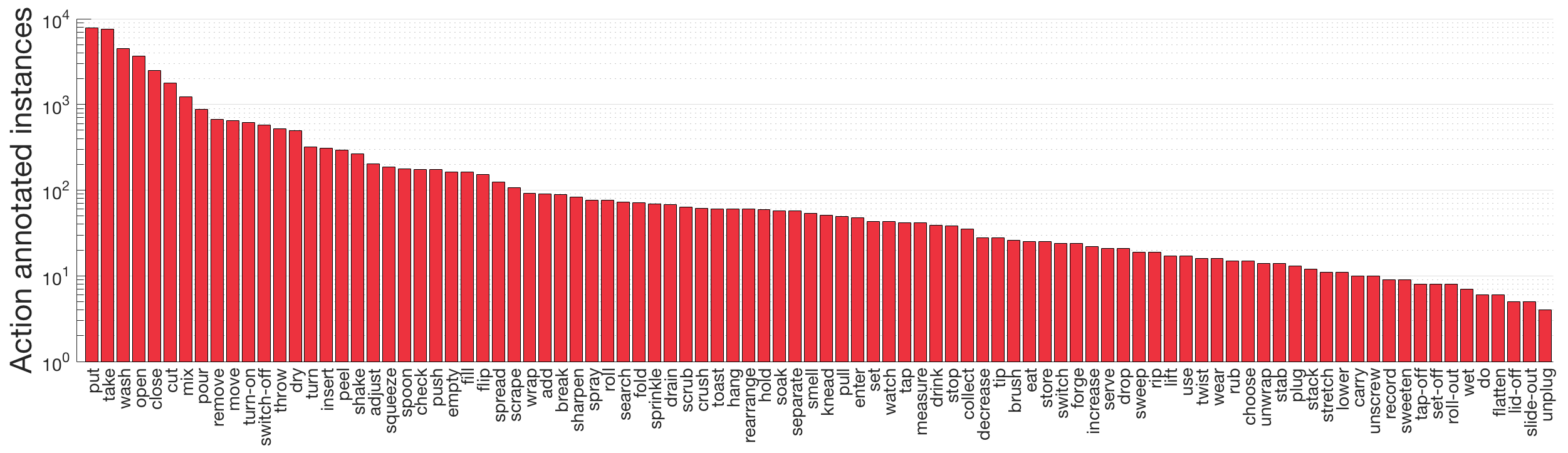}\\
\includegraphics[width=1\textwidth]{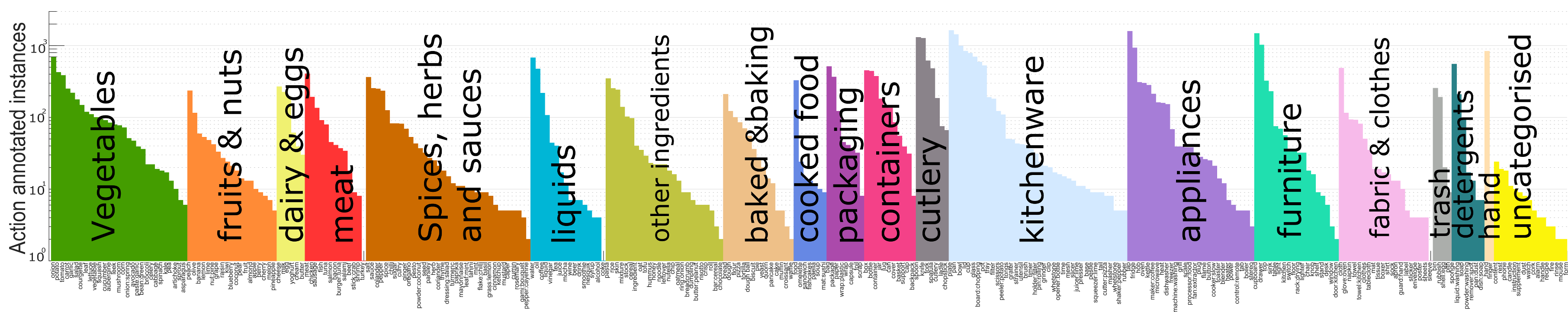}\\
\includegraphics[width=1\textwidth]{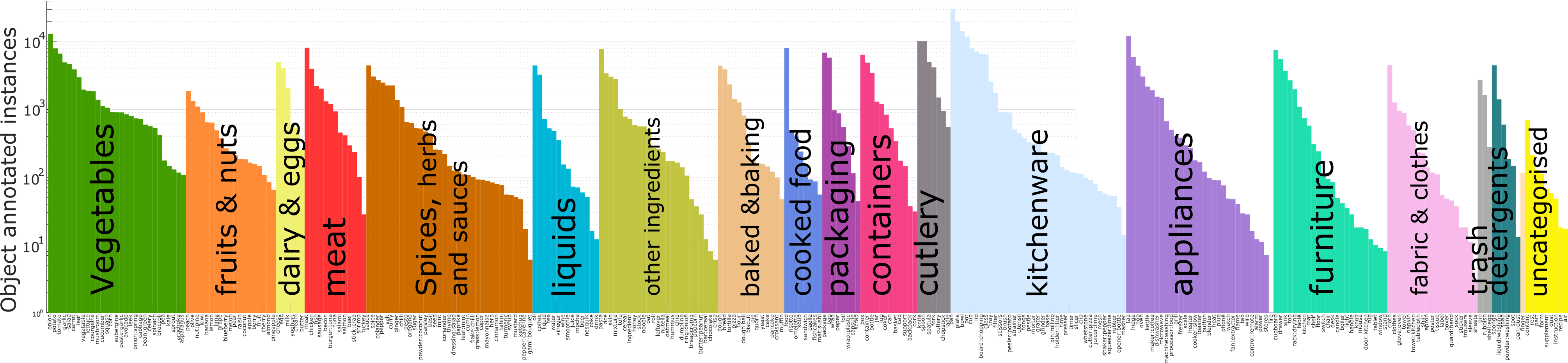}\\
\includegraphics[width=1\textwidth]{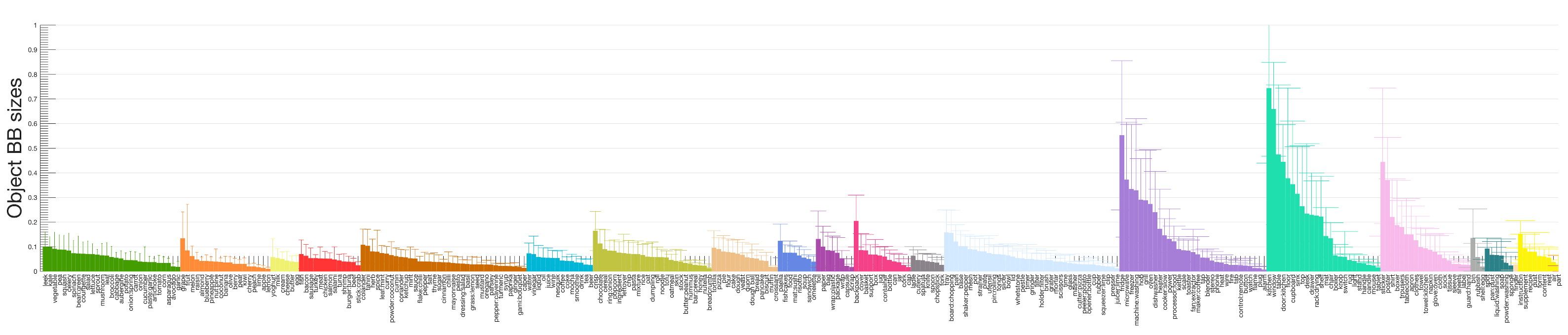}
\caption{{\bf From Top}: Frequency of verb classes in action segments; \added{Frequency of noun clusters in action segments, by category;} Frequency of noun clusters in bounding box annotations, by category; \added{Mean and standard deviation of bounding box, by category}}
\label{fig:stats}
\end{figure*}

\noindent where $\text{BB}(q,f, k)$ is the $k^{th}$ bounding box annotation by annotator $q$ in frame $f$.
Ties are broken by selecting the worker who provides the tighter bounding boxes.
\added{Figure~\ref{fig:object_ann} shows multiple annotations for five keyframes in a sequence. 
In the Figure, we distinguish between the best annotator per frame, and the overall (i.e. sequence-level) best annotator from Eq~\ref{eq:overallBestAnnotator}.
Numerically, the average IoU between consecutive frames is increased from 0.41 when the best per-frame annotator is selected, to 0.46 using our proposed approach.
Similarly the difference in the aspect ratio decreases from 0.25 to 0.22 when using the overall best annotator.
While the camera moves between frames, and the object can change size, a higher overall IoU highlights the method's consistency.}

In total, we collected 454,158 bounding boxes (per frame: $\mu~=~1.64$ boxes, ${\sigma = 0.92}$).
\added{Additionally, 125,375 true negative object labels were collected - that is the active object being absent from the image due to occlusion or camera viewpoint.
We use both annotations (bounding boxes and true negatives) in reporting results of the Object Detection challenge.}

\subsection{Verb and Noun Classes}
\label{subsec:clusters}

Since our participants annotated using free text in multiple languages, a variety of verbs and nouns have been collected.
For example, `put', `place', `put-down', `put-back', `leave' or `return' have all been used to indicate putting an object in a certain location.
We attempt to group these into classes with minimal semantic overlap, to accommodate the more typical approaches to multi-class detection and recognition where each example is believed to belong to one class only.
We estimate Part-of-Speech~(PoS), using spaCy's English core web model, to determine the verbs and nouns in the phrase.
This was necessary as although the majority of annotations are verb-noun phrases, such as `take cup' or `open fridge', there were annotations which included prepositions such as `put pan on hob'  as well as annotations which included multiple objects such as `put down onion and knife'.
We find the verb by selecting the first verb in the sentence, and find all nouns in the sentence excluding any that match the chosen verb. When a noun is absent or replaced by a pronoun (\textit{e.g. `it'}), we use the noun from the directly preceding narration (e.g. $p_i$: `rinse cup', $p_{i+1}$: `place it to dry').

\begin{table}[t!]
\centering
\caption{Sample Verb and Noun Classes. \added{Words in \textit{italics} indicate manually adjusted groupings.}}
\resizebox{1\linewidth}{!}{%
\begin{tabular}{l|l|l}
&ClassNo (Key) &Clustered Words\\
\hline
\multirow{5}{*}{\rotatebox{90}{\textbf{VERB}}} &0 (take) &take, grab, pick, get, fetch, pick-up,  collect-from, ...\\
&3 (close) &close, close-off, \textit{shut}\\
&12	(turn-on) &turn-on, start, begin, ignite, switch-on, \textit{activate}, \textit{light}, ...\\
&17 (adjust) &adjust, \textit{change}, \textit{regulate}\\
&35 (sort) &sort, rearrange, arrange, \textit{clear}, \textit{tidy}, \textit{line-up}\\\hline
\multirow{5}{*}{\rotatebox{90}{\textbf{NOUN}}} &1 (pan) &pan, frying pan, saucepan, wok, ...\\
&8 (cupboard) & cupboard, cabinet, \textit{locker}, \textit{flap}, cabinet door, closet, ...\\
&45 (garlic) &garlic, chopped garlic, garlic piece, clove, \textit{bulb}\\
&51 (cheese) &cheese, cheese slice, mozzarella, paneer, parmesan, ...\\
&78 (top) &top, counter, counter top, \textit{surface}, kitchen counter, \textit{tiles}, ...
\end{tabular}}
\label{tab:classes}
\end{table}

We refer to the set of minimally-overlapping verb classes as $C_V$, and similarly $C_N$ for nouns.
We attempted to automate the clustering of verbs and nouns using combinations of WordNet~\cite{miller1995wordnet}, Word2Vec~\cite{mikolov2013efficient}, and Lesk algorithm~\cite{Banerjee2002}, however, due to limited context there were too many meaningless clusters.
\added{Even by training Word2Vec with the PoS appended to each word, we found that both the Wikipedia corpus as well as a cookbook corpus were not suitable for our needs in clustering semantically similar words.
We tried to use WordNet for automatic clustering via assigning synsets using the simplified Lesk~\cite{lesk1986automatic,kilgarrif2000english} algorithm in order to cluster the verbs/nouns. However, we found that due to the annotations not being full sentences, in addition to consisting of only a few words, the Lesk algorithm didn't have enough context to find the correct synset.}
\revision{We automatically preprocessed the compound nouns \textit{e.g.}~`pizza cutter' as a subset of the second noun \textit{e.g.}~`cutter'.
We then manually adjusted the clustering, 
merging the variety of names used for the same object, \textit{e.g.} `cup' and `mug', as well as splitting some base nouns,
\textit{e.g.}~`washing machine' vs `coffee machine'.
We elected to manually cluster the verbs and nouns ourselves, due to the challenges of communicating this goal to annotators. This was done iteratively until authors were satisfied with the clusters}.
Table~\ref{tab:classes} shows a sample of grouped verbs and nouns into classes.
\added{We highlight a few cases (\textit{in italics}) that required manual intervention. For example, verbs `change' and `regulate' were only used in the context of heat adjustment. Similarly, `bulb' was used to refer to garlic throughout. Due to the multi-lingual nature of the narrations, words `armadietto' and `armadio' were used to refer to kitchen cupboards which were translated by AMT Workers into a `locker'. We thus group `locker' into the same class as `cupboard'.}

We provide the classes and manual clustering with the dataset. In total, we have 125 $C_V$ classes and 331 $C_N$ classes.

\begin{figure*}[t]
{\includegraphics[width=0.5\linewidth]{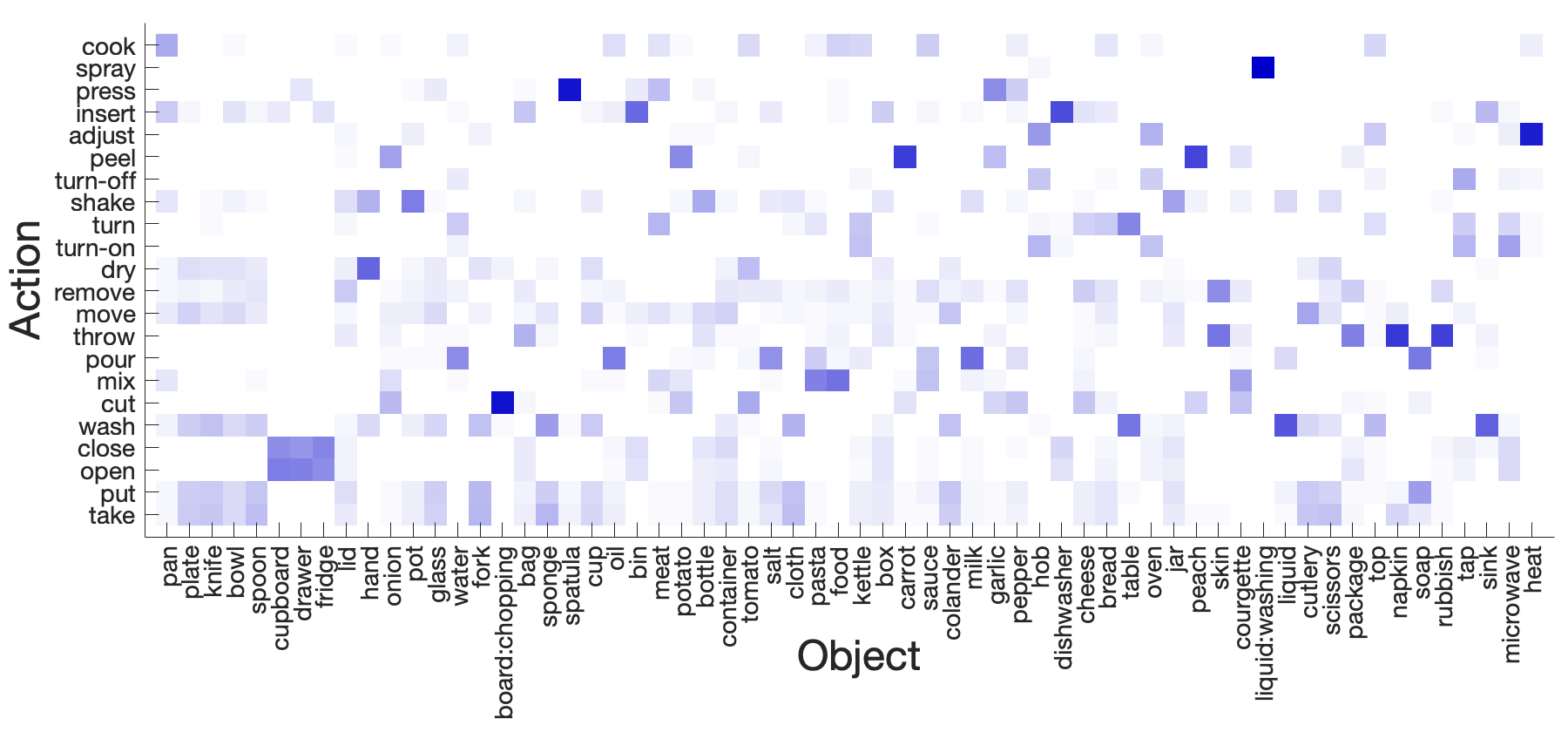}}
{\includegraphics[width=0.21\linewidth]{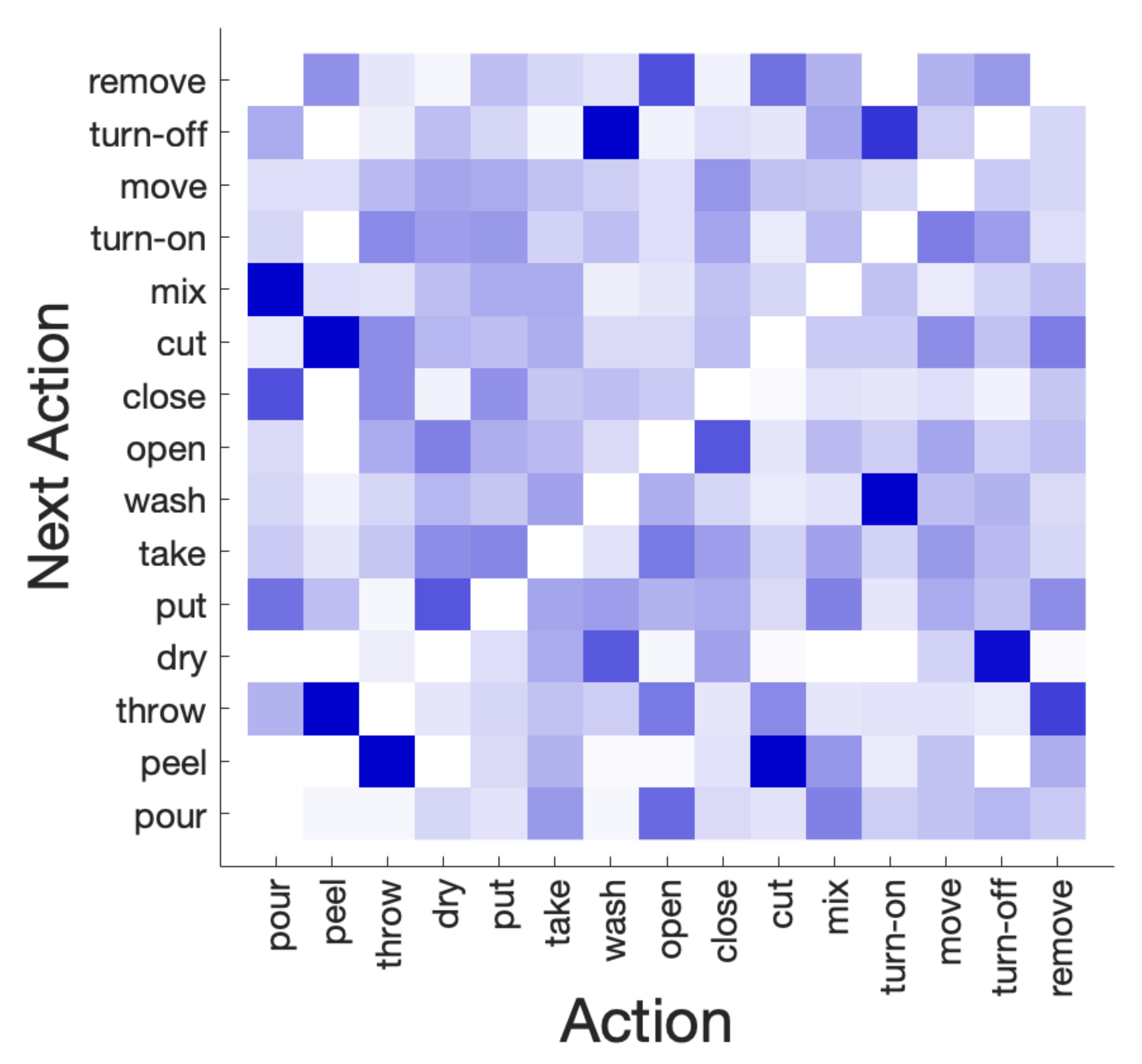}}
{\includegraphics[width=0.27\linewidth]{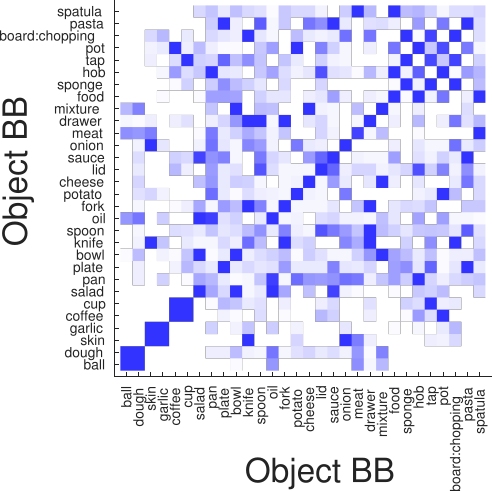}}
\caption{\small {\bf Left}: Frequently co-occurring verb/nouns in action segments [e.g. (open/close, cupboard/drawer/fridge), (peel, carrot/onion/potato/peach), (adjust, heat)]; {\bf Mid}:~Next-action excluding repetitive instances of the same action [e.g. pour $\rightarrow$ mix, peel $\rightarrow$ cut or peel $\rightarrow$ throw, turn-on $\rightarrow$ wash].; {\bf Right}: Co-occurring bounding boxes within the same annotated frame [e.g.~(cup, coffee), (knife, chopping board), (tap, sponge)]}
\label{fig:co-occur}
\end{figure*}

\begin{figure}[t]
\centering
\includegraphics[width=1\linewidth]{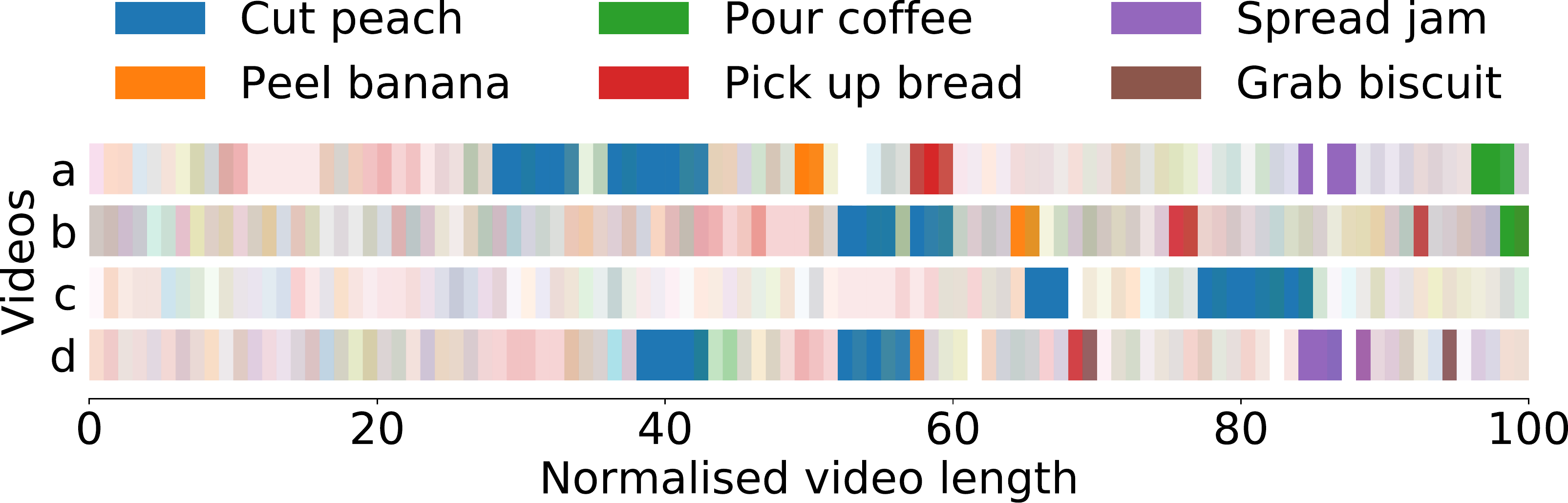}
\caption{\added{Routine activity over four days (rows), during breakfast for the same subject (P22). Coloured segments illustrate the actions performed in the video, with colours indicating action classes and gaps corresponding to background (unlabelled frames). Action segments are normalised by the video length. In total 245 action classes were present. We highlight 6 classes to demonstrate the variability in routines over several days.}}
\label{fig:routines}
\end{figure}

\subsection{Annotation Visualisations}

In Fig.~\ref{fig:stats}, we show visualisations of the annotated classes: $C_V$ ordered by frequency of occurrence in action segments, \added{$C_N$ ordered by frequency of occurrence in action segments} as well as $C_N$ ordered by number of annotated bounding boxes.
These are grouped into 19 super categories, of which 9 are food and drinks, with the rest containing kitchen essentials from appliances to cutlery.
\added{The figure also shows the sizes of the annotated bounding boxes for these categories. As objects are mostly visible at an arm's length (i.e. during usage), their sizes vary with furniture and appliances significantly larger than food ingredients.}

\begin{figure*}
\centering
\includegraphics[width=\textwidth]{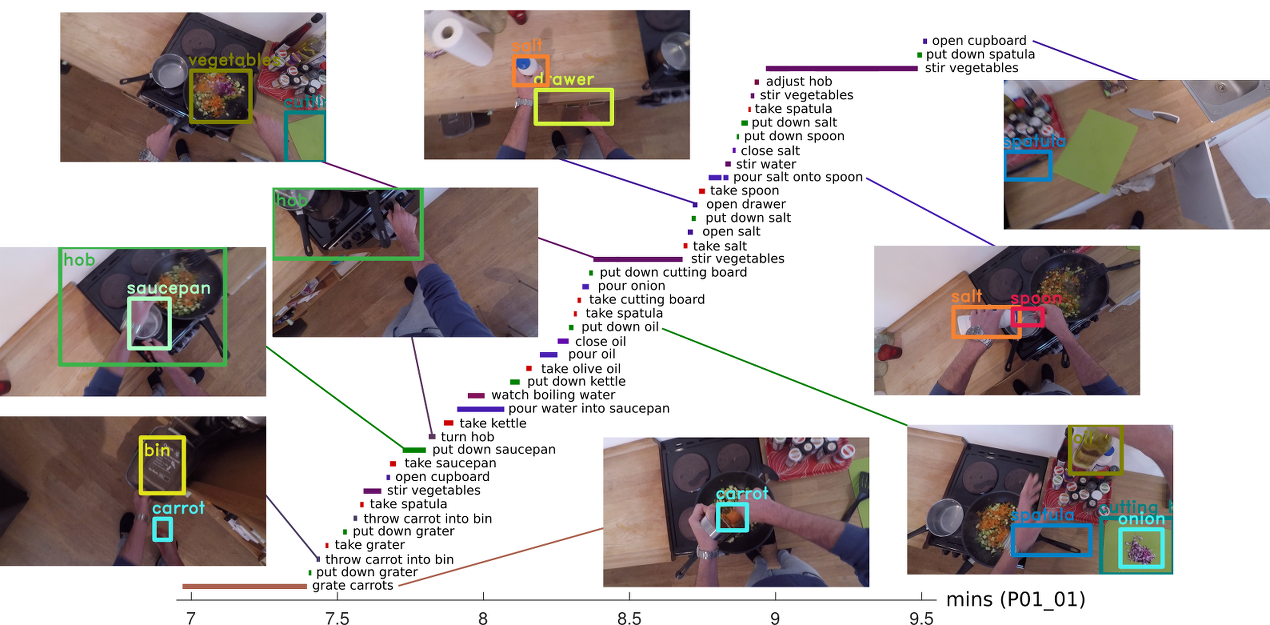}
\caption{Sample consecutive action segments with keyframe object annotations}
\label{fig:timeline_example}
\end{figure*}

Co-occurring classes are presented in Fig.~\ref{fig:co-occur}.
To demonstrate how active objects relate to the annotated segments, sample action segments and object bounding boxes are shown in Fig.~\ref{fig:timeline_example}.
\added{The figure shows how multiple active objects are annotated in frames of neighbouring actions, the variability in the sizes of bounding boxes as well as the variety of lengths of action segments.}

\added{Figure~\ref{fig:routines} depicts the activity of preparing breakfast performed by the same participant over four days. The participant prepared a similar breakfast during the four days, eating fruit (peach and banana), bread with jam and coffee. Being natural and unscripted, we observe variability in how the routine of making breakfast is carried out. For example, coffee was prepared only in two out of the fours days (videos \ref{fig:routines}a and \ref{fig:routines}b), whereas in one day (video \ref{fig:routines}c) the subject ate only peach and spent most of the time in the morning washing and tiding up the kitchen. On one day the participant treated himself to a biscuit (video \ref{fig:routines}d). White gaps in the figure correspond to unlabelled (background) frames. Importantly, there are very few background frames in the videos, which shows the density and the complexity of the dataset.}

\begin{table*}[t]
\begin{center}
\caption{Statistics of test splits: seen (\textbf{S1}) and unseen (\textbf{S2}) kitchens\revision{: number of verb, noun, action and zero-shot (ZS) classes}.}
\label{tab:s1_stats}
\resizebox{\linewidth}{!}{%
\begin{tabular}{lrrrrrrrrrrrrr}
\toprule
& \#Subjects & \#Sequences & Duration (s) & \% &Narrated &Action &Bounding & \multicolumn{2}{c}{\revision{\# Verb Classes}} & \multicolumn{2}{c}{\revision{\# Noun Classes}} & \multicolumn{2}{c}{\revision{\# Action Classes}} \\
& & & & & Segments & Segments & Boxes & \revision{Total} & \revision{ZS} & \revision{Total} & \revision{ZS} & \revision{Total} & \revision{ZS} \\
\midrule
Train/Val &28 &272 &141731 & 73\% &28,588 &28,561 &326,298 & \revision{119} & \revision{--} & \revision{321} & \revision{--} & \revision{2,513} & \revision{--} \\
\textbf{S1} Test &28 &106 &39084 &20\% &8,069 &8,064 &97,865& \revision{94} & \revision{4} & \revision{237} & \revision{15} & 
\revision{1,241} & \revision{318}\\
\textbf{S2} Test &4 &54 &13231 &7\% &2,939 &2,939 &29,995& \revision{64} & \revision{1} & \revision{139} & \revision{13} & \revision{634} & \revision{220}\\
\bottomrule
\end{tabular}}
\end{center}
\end{table*}

\subsection{Annotation Quality Assurance}
To analyse the quality of annotations, we choose 300 random samples, and manually assess correctness. We report:

\begin{itemize}[leftmargin=*]
\item \textbf{Action Segment Boundaries ($A_i$)}: We check that the start/end times fully enclose the action boundaries, with any additional frames not part of other actions - error: \textit{5.7\%}.
\item \textbf{Object Bounding Boxes ($\mathcal{O}_i$)}: We check that the bounding box encapsulates the object or its parts, with minimal overlap with other objects, and that all instances of the class in the frame have been labeled -- error: \textit{6.3\%}.
\item \textbf{Verb classes ($C_V$):} We check that the verb class is correct -- error: \textit{3.3\%}.
\item \textbf{Noun classes ($C_N$):} We check that the noun class is correct -- error : \textit{6.0\%}.

\end{itemize}

\noindent These error rates are comparable to other datasets~\cite{zhao2017slac}.

\section{Benchmarks and Baseline Results}
\label{sec:baselines}

\EPIC{} offers a variety of potential challenges from routine understanding, to activity recognition and object detection. As a start,
we define three challenges for which we provide baseline results.
For the evaluation protocols, we hold out ground truth annotations for 27\% of the data (Table~\ref{tab:s1_stats}).
We particularly aim to assess the generalizability to novel environments, and we thus structured our test set to have a collection of \textit{seen} and previously \textit{unseen} kitchens:

\noindent\textbf{Seen Kitchens (S1):} In this split,
each kitchen is seen in both training and testing, where roughly 80\% of sequences are in training and 20\% in testing. We do not split sequences, thus each sequence is in either training or testing.

\noindent\textbf{Unseen Kitchens (S2):} This divides the participants/kitchens so all sequences of the same kitchen are either in training or testing. We hold out the complete sequences for 4 participants for this testing protocol. The test set of S2 is only 7\% of the dataset in terms of frame count, but the challenges remain considerable.

\revision{In Table~\ref{tab:s1_stats}, we detail the number of classes of verbs, nouns and actions in each split. We list the number of Zero-Shot (ZS) classes in each case. For example, in \textbf{S2}, one verb and 13 nouns have not been observed in training. Additionally, 220 actions are zero-shot. These are in majority new combinations of individually observed verbs and nouns in training. For example, ``crush can'' is a zero-shot action in \textbf{S1} despite the presence of both verb (e.g.~``crush'') and noun (e.g. ``can'') separately in the training set.}

We now evaluate several existing methods on our benchmarks, to gain an understanding of how challenging our dataset is.

\begin{table*}[t!]
\centering
\caption{Baseline results for the Object Detection challenge}
\label{tab:results_obj}
\resizebox{\textwidth}{!}{%
\begin{tabular}{@{}ll|ccccccccccccccc|c|c|c@{}}

& &\multicolumn{15}{c|}{\textbf{15 Most Frequent Object Classes}} &\multicolumn{3}{c}{\textbf{Totals}}\\ \cline{3-20}
 &mAP & pan & plate & bowl & onion & tap & pot & knife & spoon & meat & food & potato & cup & pasta & cupboard & lid & few-shot & many-shot & all\\\hline
 \multirow{3}{*}{\rotatebox{90}{\textbf{S1}}} & IoU $>0.05$ & 74.00 & 72.61 & 71.50 & 60.72 & 84.44 & 69.97 & 44.03 & 40.93 & 29.65 & 58.52 & 62.82 & 53.30 & 78.39 & 51.95 & 62.77 & 9.71 & 49.80 & 38.23\\
 & IoU $>0.5$  & 67.60 & 66.21 & 65.98 & 39.96 & 73.80 & 64.71 & 28.80& 23.89 & 20.75 & 49.85 & 55.48 & 42.99 & 69.75 & 29.20 & 58.48 & 6.98 & 36.50 & 28.06\\
 & IoU $>0.75$ & 21.94 & 44.60 & 39.48 & 3.52 & 25.83 & 19.67 & 3.42 & 2.59 & 5.27 & 15.78 & 13.18 & 8.00 & 24.53 & 4.05 & 26.51 & 0.36 & 8.73 & 6.50
 \\\hline
 \multirow{3}{*}{\rotatebox{90}{\textbf{S2}}} & IoU $>0.05$ & 75.94 & 87.36 & 72.72 & 47.61 & 78.14 & 75.92 & 55.51 & 41.28 & 71.59 & 38.61 & N/A & 44.62 & 80.58 & 53.88 & 58.40 & 6.00 & 51.71 & 40.61\\
 & IoU $>0.5$ & 62.88 & 84.86 & 68.61 & 32.18 & 59.75 & 62.86 & 39.60 & 27.52 & 53.54 & 35.47 & N/A & 39.19 & 76.27 & 32.54 & 49.36 & 5.32 & 36.27 & 28.57\\
 & IoU $>0.75$  & 14.56 & 62.82 & 38.44 & 2.25 & 4.89 & 14.91 & 3.85 & 1.51 & 9.56 & 8.10 & N/A & 7.60 & 43.30 & 5.61 & 25.48 & 0.18 & 9.05 & 7.04\\ \hline
\end{tabular}
}
\end{table*}

\begin{figure*}[t]
\includegraphics[width=1\textwidth]{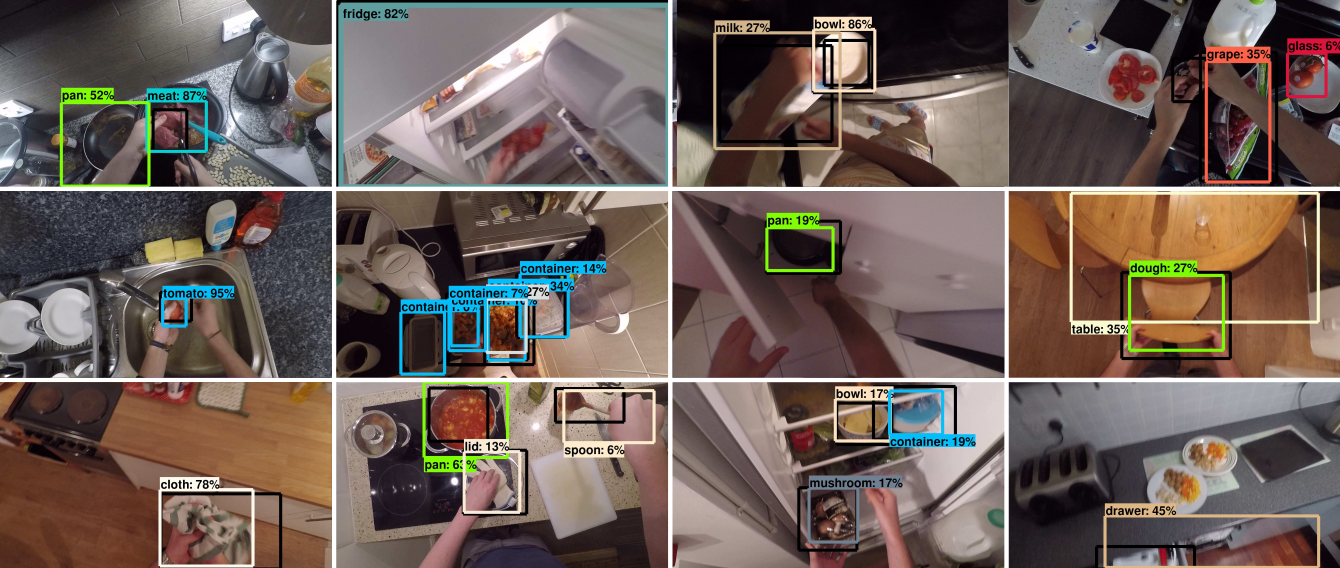}
\caption{Qualitative results for the object detection challenge. \revision{Third column shows successful detections under occlusion}. Right-hand column shows failure cases.}
\label{fig:qual_obj}
\end{figure*}

\subsection{Object Detection Benchmark}
\label{subsec:obj_challenges}
\noindent\textbf{Challenge:}
This challenge focuses on object detection for all of our $C_N$ classes.
Note that our annotations only capture the `active' objects pre-, during- and post- interaction.
We thus restrict the images evaluated per class to those where the object has been annotated.
We particularly aim to break the performance down into multi-shot and few-shot class groups, so as to analyze the capabilities of the approaches to quickly learn novel objects (with only a few examples). Our challenge leaderboard will reflect the methods' abilities on both sets of classes.

\noindent\textbf{Method:} We evaluate object detection using Faster R-CNN~\cite{ren2015faster} due to its state-of-the-art performance. \revision{It is composed of two modules that share a backbone CNN: a region proposal network~(RPN) and a detection network. The RPN outputs multi-scale class agnostic object proposals by sliding over the convolutional feature map of the last layer of the shared CNN. For each proposal the network outputs an objectness score and regresses the coordinates of a refined bounding box. The detection network functions similarly to Fast R-CNN~\cite{girshick2015fast} but considers RPN's region proposals as input. For each proposal, features from the backbone network are obtained with region of interest (ROI) pooling. The network classifies the proposals and further adjusts the bounding box coordinates. These two modules are trained alternately.
Other implementation details are in supplementary material Section B.1.
}

\noindent\textbf{Evaluation Metrics:}
We use the mean average precision (mAP) metric from PASCAL VOC~\cite{pascal}, using IoU thresholds of 0.05, 0.5 and 0.75 similar to MS-COCO~\cite{coco}. For each class, we only report results on $I^{c_n \in C_N}$, these are all images where class $c_n$ has been annotated.

\noindent\textbf{Results:} We report results in Table~\ref{tab:results_obj} for many-shot classes (those with $\geq 100$ bounding boxes in training) and few shot classes (with $\geq 10$ and $< 100$ bounding boxes in training), alongside AP for the 15 most frequent classes. There are a total of 202 many-shot classes and 78 are few-shot. One can see that our objects are generally harder to detect than in most existing datasets, with performance at the standard IoU$>0.5$ below $30\%$. Even at a very small IoU threshold, the performance is relatively low. The more challenging classes are ``meat'', ``knife'', and ``spoon'', despite being some of the most frequent ones. Notice that the performance for the low-shot regime is substantially lower than in the many-shot regime, falling short of $10\%$. This points to interesting challenges for the future. However, performances for the \emph{Seen} and \emph{Unseen} splits in object detection are comparable, thus showing generalization capability across environments.
\revision{We also assess the performance in relation to bounding box size (Table~\ref{tab:object_size}), similar to MS-COCO~\cite{coco}. We group ground-truth bounding boxes by size, into 5 categories of comparable number of instances, and report mAP for each. Performance is comparable, with best performance for the smallest objects, on both test splits.}

Figure~\ref{fig:qual_obj} shows qualitative results  with detections shown in color and ground truth shown in black.

\revision{\begin{table}[t]
\begin{center}
\caption{\revision{Object Detection mAP (IoU $>$ 0.5) the ratio of the object bounding box's area (A) to the frame}}
\label{tab:object_size}
\begin{tabular}{lrrrrr}
\toprule
A\% &0-2&2-3.5&3.5-6&6-12 &$\ge$ 12\\
\midrule
\textbf{S1} & 41.42 & 38.68 & 35.28 & 34.70& 40.20\\
\textbf{S2}  & 43.87  & 39.83 & 34.42  & 30.99 & 35.32\\
\bottomrule
\end{tabular}
\end{center}
\end{table}
}
\subsection{Action Recognition Benchmark}
\label{sec:action_recognition_benchmark}
\noindent\textbf{Challenge:} Given an action segment $A_i = [t_{s_i}, t_{e_i}]$, we aim to classify the segment into its action class, where classes are defined as ${C_a = \{(c_v \in C_V, c_n \in C_N)\}}$, and $c_n$ is the first noun in the narration when multiple nouns are present.

\begin{table*}
\begin{center}
\caption{\added{Baseline results for the action recognition challenge}}
\label{tab:results_recognition}
\resizebox{\textwidth}{!}{%
\begin{tabular}{ll|ccc|ccc|ccc|ccc}

                        &        & \multicolumn{3}{c|}{\textbf{Top-1 Accuracy}} & \multicolumn{3}{c|}{\textbf{Top-5 Accuracy}} & \multicolumn{3}{c|}{\textbf{Avg Class Precision}} &\multicolumn{3}{c}{\textbf{Avg Class Recall}} \\\cline{3-14}
                      &                        & VERB   & NOUN  & ACTION & VERB  & NOUN  & ACTION & VERB  & NOUN  & ACTION & VERB  & NOUN  & ACTION \\ \hline
\multirow{7}{*}{\rotatebox{90}{\textbf{S1}}}
                      & Random Chance          & 12.50  & 01.70 & 00.49 & 43.23 & 08.07 & 02.45 & 03.61 & 01.16 & 00.10 & 03.61 & 01.16 & 00.10 \\
                      & Largest Class          & 22.49  & 04.45 & 02.16 & 69.94 & 18.67 & 08.36 & 00.87 & 00.06 & 00.00 & 03.85 & 01.41 & 00.12 \\
                      & Time of Day            & 23.57  & 07.69 & 03.57 & 70.93 & 29.00 & 13.58 & 03.60 & 01.62 & 00.06 & 04.61 & 04.04 & 00.36 \\

                      & TSN (RGB)             & 48.02  & 38.98 & 22.44 & 87.00 & 65.75 & 44.91 & 51.08 & 37.04 & 13.64 & 28.83 & \textbf{34.42} & 11.82 \\
                      & TSN (Flow)            & 52.37  & 26.95 & 16.84 & 84.81 & 50.84 & 34.19 & 43.87 & 24.28 & 07.66  & 24.81 & 18.99 & 05.54  \\
                    & TSN (Audio) & 41.05 & 20.48 & 12.26 & 79.36 & 43.46 & 26.75 & 27.27 & 17.56 & 05.73 & 22.42 & 17.16 & 05.02\\
                      & TSN (RGB+Flow+Audio) & \textbf{56.10} & \textbf{40.28} & \textbf{26.06} & \textbf{87.69} & \textbf{66.33} & \textbf{48.18} & \textbf{58.45} & \textbf{41.09} & \textbf{15.45} & \textbf{29.42} & 34.07 & \textbf{12.43}\\ \hline
\multirow{7}{*}{\rotatebox{90}{\textbf{S2}}}
                      & Random Chance          & 10.62  & 01.90 & 00.58 & 38.70 & 09.08 & 02.89 & 03.51 & 01.06 & 00.08 & 03.51 & 01.06 & 00.08 \\
                      & Largest Class          & 22.26  & 04.71 & 02.59 & 63.43 & 19.12 & 08.95 & 00.86 & 00.07 & 00.00 & 03.85 & 01.41 & 00.12 \\
                      & Time of Day            & 23.52  & 10.99 & 05.36 & 69.00 & 35.13 & 18.20 & 03.12 & 01.56 & 00.06 & 04.59 & 06.12 & 00.38 \\

                      & TSN (RGB)            & 36.50  & 22.91 & 11.30 & 74.74 & 47.25 & 26.60 & 22.20 & 16.57 & 07.09  & 12.86 & 17.65 & 07.12  \\
                      & TSN (Flow)           & 47.49  & 21.75 & 14.20 & 76.85 & 42.95 & 28.03 & \textbf{25.94} & 14.24 & 06.26  & \textbf{19.50} & 15.59 & 07.20  \\
                      & TSN (Audio) & 37.05 & 11.31 & 05.95 & 68.56 & 30.62 & 16.02 & 18.72 & 09.88 & 03.69 & 17.86 & 09.46 & 03.44\\
                      & TSN (RGB+Flow+Audio) & \textbf{48.42} & \textbf{25.20} & \textbf{15.61} & \textbf{77.98} & \textbf{50.43} & \textbf{31.79} & 24.29 & \textbf{18.24} & \textbf{08.35} & 18.29 & \textbf{18.99} & \textbf{09.06}\\
\end{tabular}}

\end{center}
\end{table*}

\begin{table*}
  \begin{center}
    \caption{\added{Action Recognition Challenge - Comparison of Temporal Modelling Architectures}}
    \label{tab:results_recognition_temporal_architectures}
    \resizebox{\textwidth}{!}{%
      \begin{tabular}{ll|ccc|ccc|ccc|ccc}
        & & \multicolumn{3}{c|}{\textbf{Top-1 Accuracy}} & \multicolumn{3}{c|}{\textbf{Top-5 Accuracy}} & \multicolumn{3}{c|}{\textbf{Avg Class Precision}} &\multicolumn{3}{c}{\textbf{Avg Class Recall}} \\\cline{3-14}
        
        &             & VERB  & NOUN  & ACTION & VERB  & NOUN  & ACTION & VERB  & NOUN  & ACTION & VERB  & NOUN  & ACTION \\ \hline
        \multirow{5}{*}{\rotatebox{90}{\textbf{S1}}}& 2SCNN & 42.08 & 30.09 & 13.30  & 81.20 & 55.05 & 31.04  & 37.97 & 30.44 & 6.01   & 14.80 & 22.06 & 5.10   \\
        & TSN         & 54.70 & 40.11 & 25.43  & 87.24 & 65.81 & 45.69  & 57.22 & 38.34 & 15.14  & 29.52 & 34.23 & 12.65  \\
        & TRN         & 61.12 & 39.28 & 27.86  & 87.71 & 64.36 & 47.56  & 52.32 & 35.68 & 16.38  & 32.93 & 34.18 & 14.36  \\
        & TRN*        & \textbf{62.68} & 39.82 & 29.41  & 87.96 & 64.94 & 48.91  & 53.19 & 35.85 & 16.75  & 34.31 & 34.19 & 14.46  \\
        & TSM         & 62.37 & \textbf{41.88} & \textbf{29.90}  & \textbf{88.55} & \textbf{66.43} & \textbf{49.81}  & \textbf{59.51} & \textbf{39.50} & \textbf{18.38}  & \textbf{34.44} & \textbf{36.04} & \textbf{15.80}  \\
        \hline
        \multirow{5}{*}{\rotatebox{90}{\textbf{S2}}}
        & 2SCNN & 35.78 & 19.22 & 8.06   & 72.58 & 40.32 & 19.63  & 18.52 & 16.54 & 4.09   & 10.88 & 13.60 & 3.79   \\
        & TSN         & 46.06 & 24.27 & 14.78  & 76.65 & 49.27 & 29.81  & 21.62 & 16.90 & 7.31   & 17.59 & 18.01 & 9.22   \\
        & TRN         & 51.62 & \textbf{26.02} & 17.34  & 78.42 & 48.99 & 32.57  & \textbf{32.47} & \textbf{19.99} & 9.45   & 21.63 & 21.53 & 10.11  \\
        & TRN*        & \textbf{52.03} & 25.88 & \textbf{17.86}  & 78.90 & 49.03 & 32.54  & 29.30 & 19.25 & \textbf{10.46}  & \textbf{21.73} & 22.23 & 11.14  \\
        & TSM         & 51.96 & 25.61 & 17.38  & \textbf{79.21} & \textbf{49.47} & \textbf{32.67}  & 27.43 & 17.63 & 9.17   & 20.19 & \textbf{22.93} & \textbf{11.18}  \\
      \end{tabular}}
  \end{center}
\end{table*}

\begin{figure*}[t!]
\includegraphics[width=\textwidth]{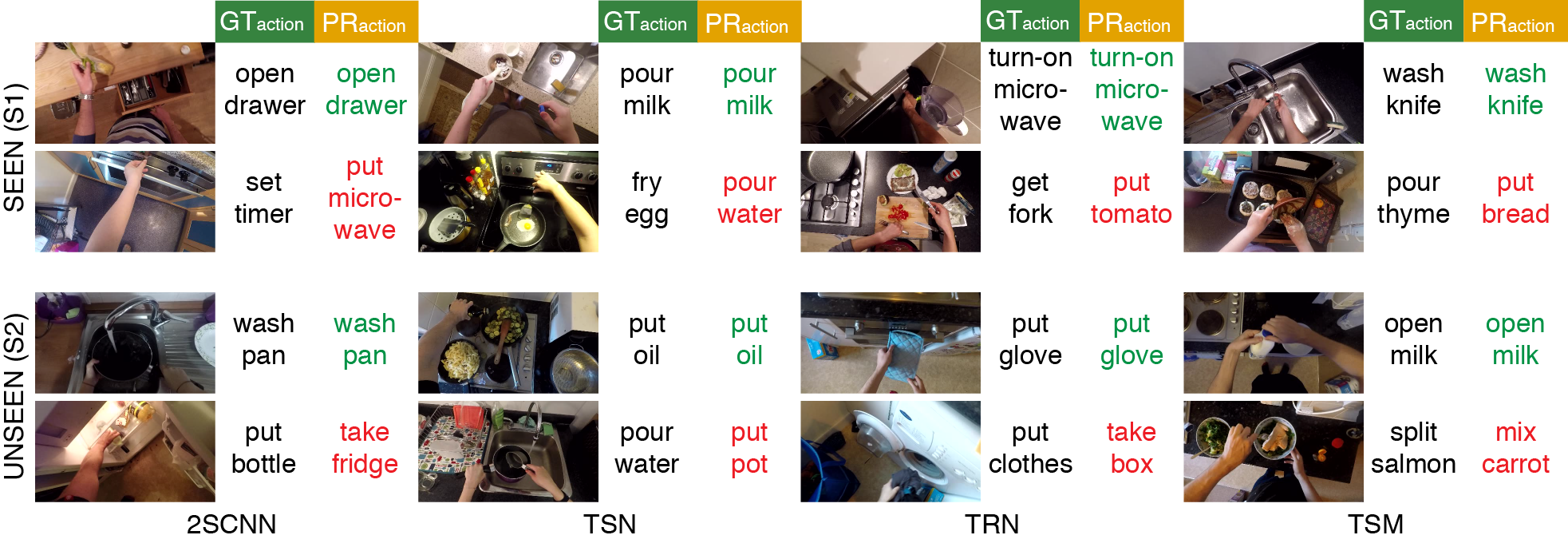}
\caption{\added{Qualitative results for the action recognition challenge. \emph{Seen} (top) and \emph{Unseen} (bottom) test sets.}}
\label{fig:recognition_qualitative}
\end{figure*}

\noindent\textbf{Evaluation Metrics:}
  We report two sets of metrics: aggregate and per-class, which are equivalent to the class-agnostic and class-aware metrics in~\cite{zhao2017slac}.
  For aggregate metrics, we compute top-1 and top-5 accuracy for correct predictions of $c_v$, $c_n$ and their combination $(c_v, c_n)$ -- we refer to these as `verb', `noun' and `action'. Accuracy is reported on the full test set. 
  For per-class metrics, we compute precision and recall for classes
  with more than 100 samples in training, then average the metrics
  across classes, these are 27 verb classes and 67 noun classes.
  We also report per-class metrics for the valid combinations of these classes (758 action classes).
  Per-class metrics for smaller classes are $\approx 0$ as %TSN
  \revision{CNN}s are better suited for classes with sufficient training data.

\noindent\textbf{Baselines:}
\added{
Three baselines are provided: \textit{Random Chance}, \textit{Largest Class}, and \textit{Time of Day}.
The Time of Day baseline groups action segments by their starting hour (Fig.~\ref{fig:statsCollect}), for a given test set, then uses the majority class within the time of day as a baseline prediction.
The Time of Day majority class baseline outperforms the baseline that does not consider the time of day knowledge, highlighting the potential benefit for daily routine modelling in the dataset.
In all baselines, action predictions are computed from the ground-truth action labels rather than naively combining verb and noun baselines.
}

\noindent\textbf{Temporal Network Architectures:}
\added{\noindent We benchmark four models to assess the need for temporal modelling in the dataset: Two-stream CNN (2SCNN)~\revision{\cite{simonyan2014twostream}}, Temporal Segment Networks~(TSN)~\cite{wang2016tsn}, Temporal Relational Networks (TRN)~\cite{zhou2018temporal}, and Temporal Shift Module (TSM)~\cite{lin2018_TSMTemporalShift}.
We summarise the similarities/differences in these architectures next.}

\revision{Inputs to all the models, \textit{snippets}, are sampled according to the TSN sampling strategy, which is the same sampling strategy used by TRN and TSM. The action segment is split into $n$ equally sized non-overlapping sub-segments and a snippet is sampled at a random position within each of these.
In a 2SCNN, one snippet is sampled randomly from the full segment (i.e. $n = 1$)
whereas for TSN, TRN and TSM, $n > 1$.
In TSN~\cite{wang2016tsn}, each snippet is propagated through a 2D CNN backbone and we aggregate the class scores across snippets through average pooling.
As a consequence, TSN observes more snippets within the segment compared to 2SCNN, but is unable to learn temporal correlations across snippets.
TRN and TSM can be viewed as evolutionary descendants of TSN, integrating temporal modelling.
In TRN~\cite{zhou2018temporal}, relations over ordered sets of snippets features of size 2 to $n$ are computed.
The relational features are fed to separate MLP classifiers at each scale and the predicted class scores are aggregated through summation.
In TSM~\cite{lin2018_TSMTemporalShift}, the CNN backbone is modified to support reasoning across segments by shifting a proportion of the filter responses across the temporal dimension.
This opens the possibility for subsequent convolutional layers to learn temporal correlations. Both TRN and TSM thus explicitly model temporal relationships within the action segment.}

\added{
We compare the models on the backbone that is part of the respective model's published code: BN-Inception~\cite{szegedy2015going} for 2SCNN, TSN and TRN versus ResNet-50~\cite{resnet} for TSM. For comparability, we also evaluate TRN using ResNet-50 backbone, which we refer to as TRN* in table and results.
All models are trained with an averaged softmax cross-entropy loss over each classification layer: ${\mathcal{L} = 0.5 (\mathcal{L}_n + \mathcal{L}_v)}$.
We obtain action predictions from verb and noun predictions assuming the tasks are independent}, as in~\cite{kalogeiton2017}.
\revision{Other implementation details to replicate these results are in supplementary material Section B.2.}

\revision{Note that none of the baselines used utilise the object bounding box annotations presented in Section~\ref{subsec:strongObject}. Those have only been used in Section~\ref{subsec:obj_challenges} for the object detection challenge.
However, a surge of recent approaches have used the object bounding boxes for action recognition on our dataset~\cite{suris2019learning,wangsymbiotic, kapidis2019egocentric}.}

\begin{figure*}[t]
\includegraphics[width=\textwidth]{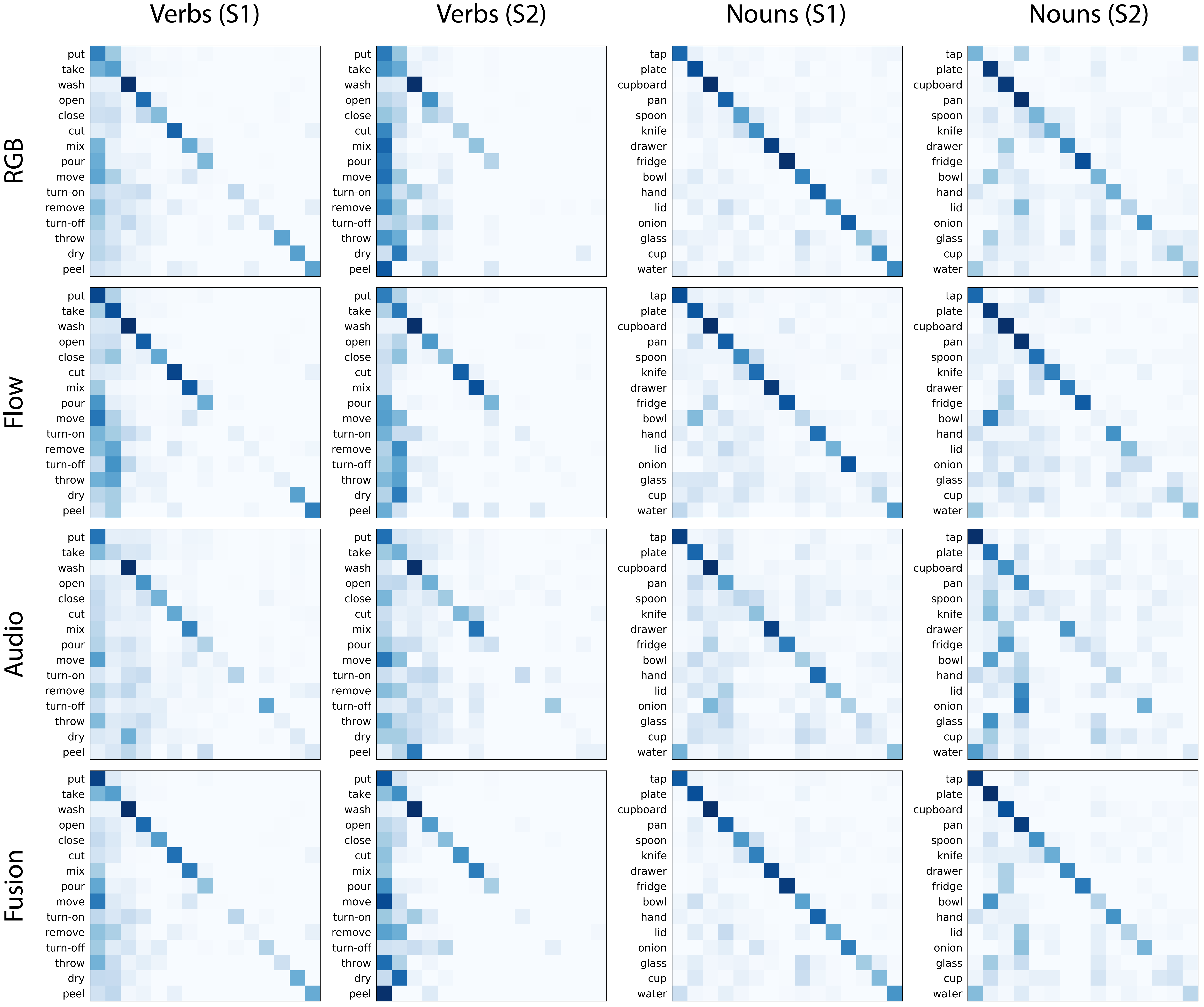}
\caption{Confusion matrices of predicted verbs and nouns for \textbf{S1} and \textbf{S2}, for each modality and late fusion of modalities.}
\label{fig:confusion_modalities}
\end{figure*}

\noindent\textbf{Modality and Fusion Results:}
\added{In Table~\ref{tab:results_recognition}, we present results of the Temporal Segment Network (TSN) on the three modalities separately - RGB, Flow and Audio, as well as their fusion. 
\revision{Note that `Audio' refers to the stereo audio stream captured by the GoPro wearable camera, alongside the video. This captures momentary sounds for some actions, e.g. putting a cup down, as well as extended sounds for others, e.g. frying or washing. Other irrelevant background sounds like music playing are also present in the dataset.}
Flow achieves best performance for verb classification while RGB gives the highest performance on noun classification. Interestingly, audio achieves good performance in both verb and noun classification, particularly on the seen (\textbf{S1}) test set. However, on the unseen test set, noun classification performance drops for both RGB and audio, making flow the most transferable modality to novel environments. Interestingly, audio outperforms RGB on the unseen (\textbf{S2}) test set for top-1 verb accuracy. On all metrics, fused results outperform or are comparable to the best individual modality. The challenge of getting both verb and noun labels correct remains significant for both \textit{seen} (top-1 accuracy 26.1\%) and \textit{unseen} (top-1 accuracy 15.6\%) environments.}

\added{We further explore the multi-modal nature of the dataset in Fig.~\ref{fig:confusion_modalities} through confusion matrices of the top-15 verb and noun classes. It is clear that confusion decreases overall with fusion.
Audio performs well for actions with sounds, e.g. `wash', and for interacting objects, e.g. `drawer' where a sound is audible when the drawer closes. As appearance is more important for nouns, flow shows more confusion in some nouns than RGB, e.g. `bowl' is confused with `plate'. Finally, the discriminative power of flow on verbs in comparison to RGB is more apparent in some cases than others, e.g. `cut' and `mix'. }
 \revision{More details on the potentials and architectures for fusing audio for egocentric action recognition in \EPIC{} have recently been noted~\cite{kazakos19iccv,Xiao20audiovisualSlowFast}.}

\begin{table*}[t]
\centering
\caption{\added{Baseline results for the action anticipation challenge}}
\label{tab:results_anticipation}
\resizebox{\linewidth}{!}{%
\begin{tabular}{ll|ccc|ccc|ccc|ccc}
&        & \multicolumn{3}{c|}{\textbf{Top-1 Accuracy}} & \multicolumn{3}{c|}{\textbf{Top-5 Accuracy}} & \multicolumn{3}{c|}{\textbf{Avg Class Precision}} &\multicolumn{3}{c}{\textbf{Avg Class Recall}} \\\cline{3-14}
&        & VERB      & NOUN      & ACTION     & VERB      & NOUN      & ACTION     & VERB   & NOUN  & ACTION & VERB   & NOUN  & ACTION \\\hline
\multirow{5}{*}{\rotatebox{90}{\textbf{S1}}}
& 2SCNN & 30.62 & 14.71 & 04.37 & 76.04 & 39.51 & 15.55 & 15.07 & 18.80 & 02.11 & 07.29 & 08.79 & 01.28\\
& TSN & \bf 32.48 & \bf 16.74 & 07.49 & \bf 76.79 & \bf 41.20 & \bf 18.95 & \bf 35.22 & \bf 22.05 & 03.92 & 08.43 & 12.50 & 03.27\\
& TRN & 31.40 & 15.82 & 07.28 & 76.74 & 40.04 & 18.84 & 27.38 & 16.74 & 03.92 & 10.21 & 12.51 & 03.11\\
& DMR & 26.53 & 10.43 & 01.27 & 73.30 & 28.86 & 07.17 & 06.13 & 04.67 & 00.33 & 05.22 & 05.59 & 00.47\\
& ED & 29.35 & 16.07 & \bf 08.08 & 74.49 & 38.83 & 18.19 & 18.08 & 16.37 & \bf 05.69 & \bf 13.58 & \bf 14.62 & \bf 04.33\\
\hline

\multirow{5}{*}{\rotatebox{90}{\textbf{S2}}}
& 2SCNN & \bf 25.61 & \bf 09.76 & 02.08 & \bf 68.79 & \bf 27.59 & 10.17 & \bf 14.37 & 05.52 & 01.32 & 05.84 & 05.88 & 00.85\\
& TSN & 24.89 & 09.63 & 02.94 & 67.87 & 26.02 & 09.49 & 12.28 & \bf 06.72 & \bf 01.53 & 05.74 & 06.07 & 01.31\\
& TRN & 24.14 & 09.05 & \bf 03.62 & 67.22 & 26.05 & \bf 10.38 & 10.82 & 06.12 & 01.39 & 06.05 & \bf 06.13 & \bf 02.06\\
& DMR & 24.79 & 08.12 & 00.55 & 64.76 & 20.19 & 04.39 & 09.18 & 01.15 & 00.55 & 05.39 & 04.03 & 00.20\\
& ED & 22.52 & 07.81 & 02.65 & 62.65 & 21.42 & 07.57 & 07.91 & 05.77 & 01.35 & \bf 06.67 & 05.63 & 01.38\\
\end{tabular}}
\end{table*}

\noindent \textbf{Temporal Model Results:}
\added{We then benchmark several models that explicitly capture temporal progression for action recognition in their architectural design.
We restrict this analysis to two modalities: RGB and Flow, in line with previous results that use these temporal models.
We report fusion results (RGB+Flow) in each case.
These models have successfully distinguished themselves on video datasets that require explicit temporal modelling compared to other datasets that can be adequately modelled from single images, or even a re-ordered set of frames. In Table~\ref{tab:results_recognition_temporal_architectures}, results indicate clear improvement when temporal modelling is incorporated for verb classes - more specifically 6.4-8.0\% for top-1 verb accuracy in \textbf{S1}, as well as 5.6-6.0\% in \textbf{S2}. Temporal modelling has understandably less of an effect on noun classification. Interestingly, TSM (ResNet-50 backbone) seems to outperform TRN (BNInception backbone) and TRN* (ResNet-50 backbone) on 11 out of 12 metrics in the seen test set, however this is not the case in the unseen test set where TRN achieves higher performance on 7 out of the 12 metrics. This could indicate the temporal shift architecture is less generalisable to unseen actions, both verbs and nouns.}

\added{\figurename~\ref{fig:recognition_qualitative} reports qualitative results, with success highlighted in green, and failures in red. Some models perform better than others, as they have better capacity for temporal modelling, e.g.~while TSM predicts `open bottle' correctly (an action with fine-grained motion), 2SCNN confuses `put bottle' with `take fridge'. A major factor for failure is object ambiguity, i.e. when multiple objects appear in the scene, for example `get fork' is predicted as
`put tomato' because both the fork and tomatoes appear in the video, and similarly `put clothes' is predicted as `take box'. Finally, appearance and motion similarity between objects and actions respectively can also cause misclassifications, where `split salmon' is predicted as `mix carrot'.}

\subsection{Action Anticipation Benchmark}
Anticipating what is going to happen in the near future is a~fundamental ability both for humans and intelligent systems. In the context of first-person vision, anticipating the wearer's next actions could be useful to automatically control smart home appliances, provide guidance on their use and notify the occurrence of possibly dangerous actions. Previous works have investigated different anticipation tasks from an egocentric perspective, including predicting future localisation~\cite{park2016egocentric}, next-active objects~\cite{furnari2017next} or forecasting actions~\cite{rhinehart2017first}.

\noindent\textbf{Challenge:}
We propose the task of egocentric action anticipation, i.e. the prediction of an action before it happens.
Specifically, let $\tau_a$ be the `anticipation time', i.e. how far in advance the action has to be recognised, and let $\tau_o$ be the `observation time', i.e. the length of the observed video segment preceding the action.
Given an action segment $A_i=[t_{s_i},t_{e_i}]$, the task is to predict the corresponding action class $C_a$ by observing the video segment \emph{preceding} the action start time $t_{s_i}$ by $\tau_a$ seconds, that is the segment of temporal bounds $[t_{s_i}-(\tau_a+\tau_o),t_{s_i}-\tau_a]$.
Results for this challenge are evaluated using the same evaluation measures discussed for the action recognition challenge (see Sec.~\ref{sec:action_recognition_benchmark}).

\noindent\textbf{Methods:}
We compare the performance of $5$ different models on the considered task.
\revision{Three models: 2SCNN~\cite{simonyan2014twostream}, TSN~\cite{wang2016tsn} and TRN~\cite{zhou2018temporal} are the action recognition approaches discussed in Section~\ref{sec:action_recognition_benchmark}, adapted for the task of action anticipation by feeding them with video segments \textit{preceding} the start of the action by $\tau_a$. The Deep Multimodal Regressor (DMR)~\cite{vondrick2016anticipating}, and the Encoder-Decoder (ED) architecture inspired by~\cite{gao2017red} are methods originally proposed for action anticipation in the context of third person videos. DMR~\cite{vondrick2016anticipating} trains a multimodal CNN to predict $K$ possible representations of a frame appearing in $\tau_a$ seconds from the observation of the current frame. The model is first trained in an unsupervised way then fine-tuned to predict future actions from these anticipated representations. ED~\cite{gao2017red} consists in an encoder-decoder LSTM which processes past frames and predicts the representations of future frames. Future actions are predicted from the anticipated representations. Similarly to DMR, ED is also pre-trained in an unsupervised way.}
All methods have been modified to predict verb and noun classes jointly as in~\cite{kalogeiton2017}, considering a fixed anticipation time of $\tau_a=1s$ as done in previous works~\cite{gao2017red,vondrick2016anticipating}. \revision{Implementation details are in the supplementary material B.3.}

\noindent\textbf{Results:}
\tablename~\ref{tab:results_anticipation} reports the results of all methods on \textbf{S1} and \textbf{S2}. For 2SCNN, TSN and TRN, we report the results obtained when $\tau_o=0.5s$. Results for other observation times are discussed later in this section. As expected, the results obtained on the action anticipation task are lower than the ones obtained in the case of action recognition. This can be assessed by comparing \tablename~\ref{tab:results_anticipation} and \tablename~\ref{tab:results_recognition}. It is worth noting that on both \textbf{S1} and \textbf{S2}, there is not a clear top-performing method across the different evaluation measures. While ED has an advantage on action-related measures in \textbf{S1}, such advantage disappears in \textbf{S2}. Despite being designed for action anticipation, DMR does not obtain good results on both sets. This is probably due to the fact that approaches based on the regression of future representations are less suited for egocentric action anticipation, due to the much higher variability exhibited by first person videos.

\begin{figure*}[t]
    \centering
    \includegraphics[width=1\linewidth]{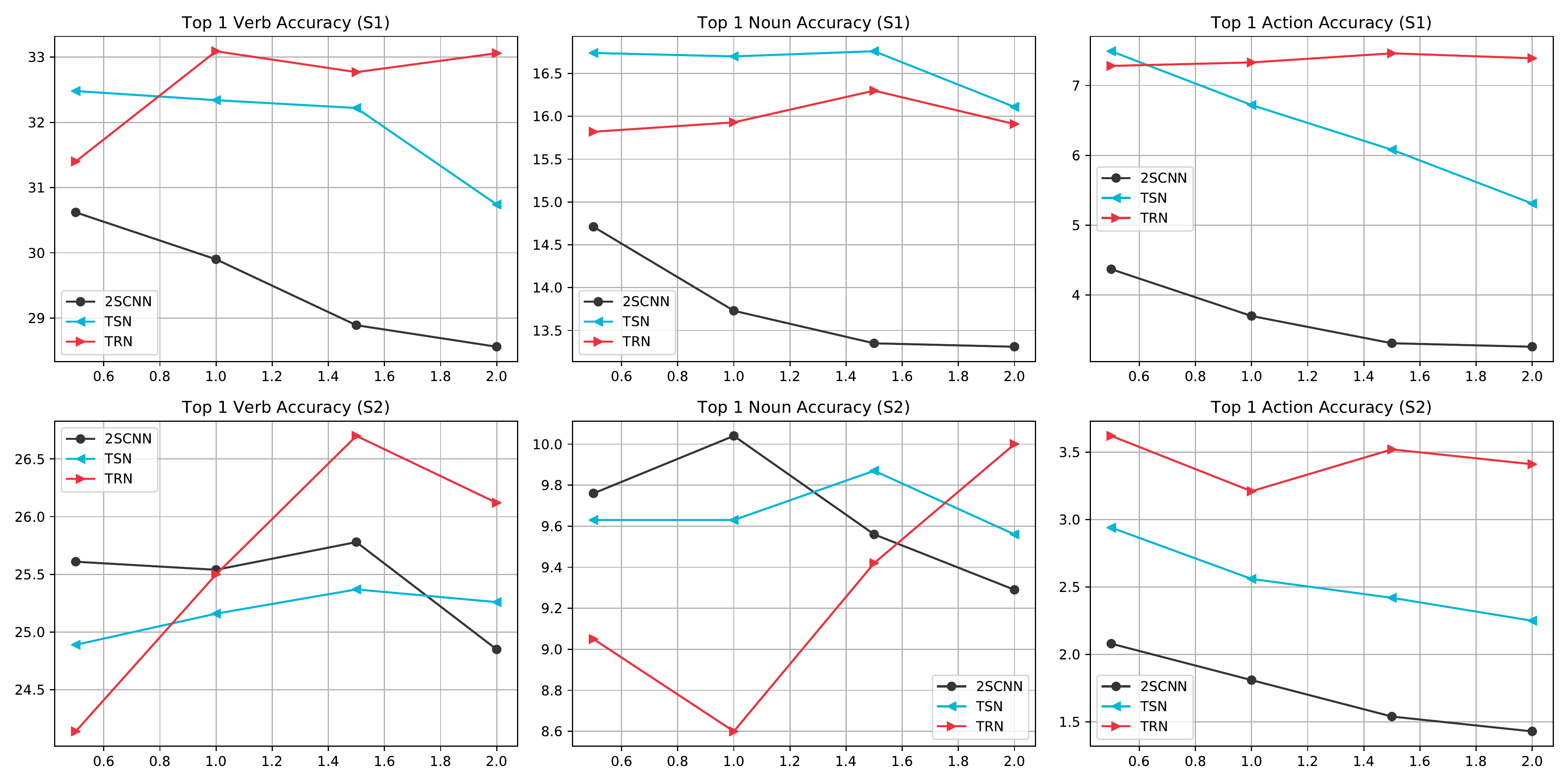}
    \caption{\added{As the observation time increases (x-axis) from 0.5 to 2.0 seconds, top-1 Verb, Noun and Action accuracy of 2SCNN, TSN and TRN on \textbf{S1} and \textbf{S2}.}}
    \label{fig:at}
\end{figure*}
\begin{figure*}[t]
\centering
\includegraphics[width=0.49\textwidth]{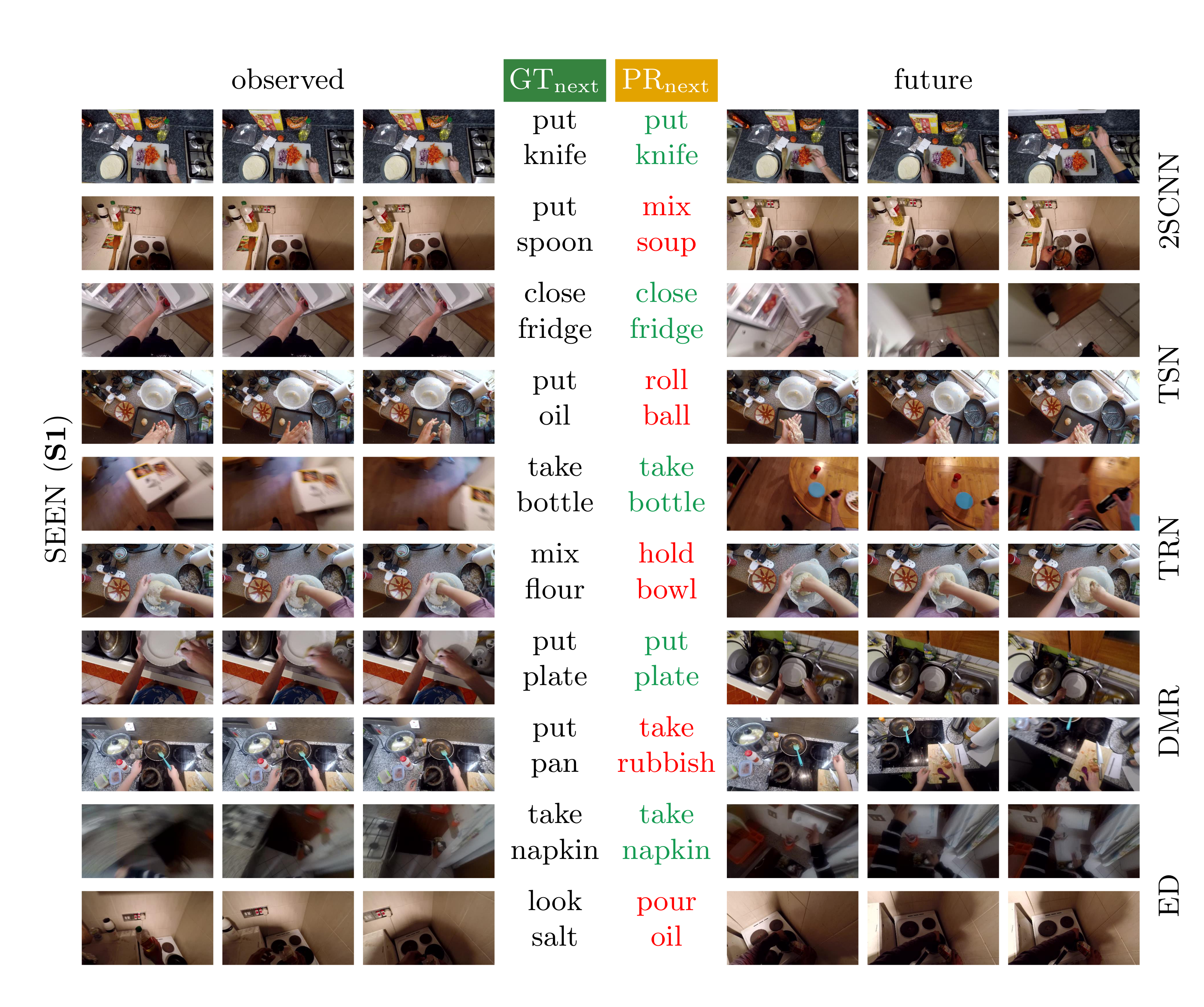}
\includegraphics[width=0.49\textwidth]{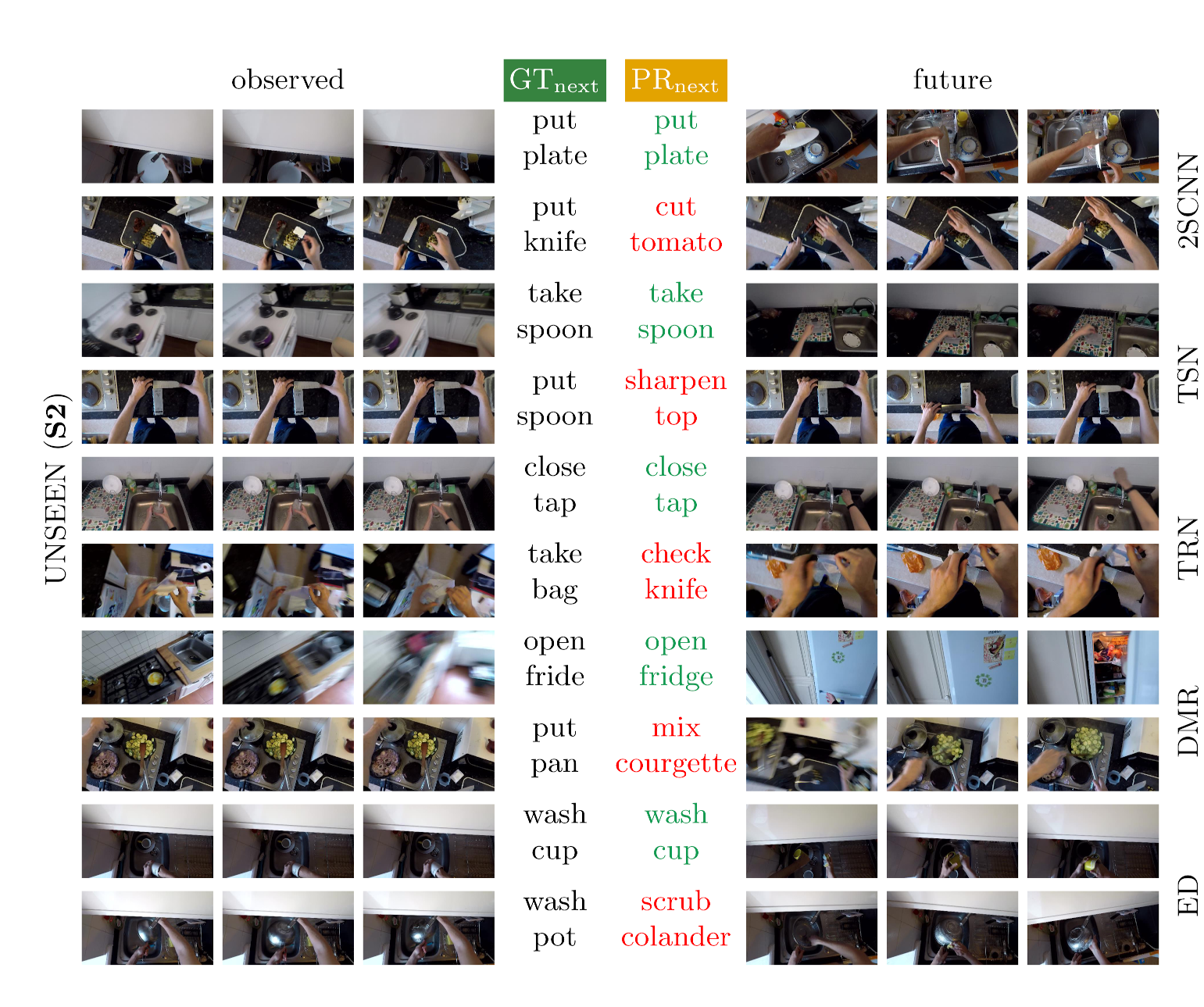}
\caption{\added{Qualitative results for the action anticipation challenge - Seen (left) and Unseen (right) test sets.}}
\label{fig:anticipation_qualitative_seen}
\end{figure*}

\added{\figurename~\ref{fig:at} reports the Top-1 Verb, Noun and Action accuracy of 2SCNN, TSN and TRN on both \textbf{S1} and \textbf{S2} when varying the observation time $\tau_o \in [0.5, 1, 1.5, 2]s$. The diagram shows how 2SCNN and TSN struggle in exploiting longer observation times, especially for action-based measures (last column). This is probably due to the limited number of frames such methods rely on during training. Indeed, with a longer observation time, most of the frames can be sampled far away from the beginning of the action, which makes anticipation harder. A different behaviour is observed for TRN, which uses a larger number of segments, and aggregates them in a more principled way. TRN is less sensitive to the choice of the observation time in \textbf{S1}. In \textbf{S2} instead, it seems to benefit more from longer observation times (see the peaks in the bottom right plots of \figurename~\ref{fig:at}). In general, it can be observed that all methods tend to be sensitive in different ways to the choice of the observation time, which suggests the limited ability of such baseline to reason about the past and the future.}

\added{\figurename~\ref{fig:anticipation_qualitative_seen} reports qualitative results. Success cases generally involve the presence in the scene of the object to be anticipated, e.g., ``put knife'' (first row left), ``wash cup'' (penultimate row right). Some failure cases are due to the inherent ambiguity of the future, e.g., ``put oil'' anticipated instead of ``roll ball'' (fourth row left), ``take bag'' anticipated instead of ``check knife'' (sixth row right). The anticipation of some actions (e.g., ``take rubbish'' in eighth row left) would require a better analysis of a longer temporal context, in which current methods are limited.}

\section{Conclusion and Future Work}
We present the largest and most varied dataset in egocentric vision to date, \EPIC{}, captured in participants' native environments. We collect 55 hours of video data recorded on a head-mounted Go-Pro, and annotate it with narrations, action segments and object annotations using a pipeline that starts with live commentary of recorded videos by the participants themselves.
Baseline results on object detection, action recognition and anticipation challenges show the great potential of the dataset for pushing approaches that target fine-grained video understanding to new frontiers. Results show the importance of temporal modelling in this dataset and the need for novel approaches to tackle the anticipation challenge.
The three defined challenges form the base for higher-level understanding of the wearer's goals. We have shown that existing methods are still far from tackling these tasks with high precision, pointing to exciting future directions. 

\revision{The unscripted and untrimmed nature of recordings in this dataset allows other explorations beyond the challenges reported in this paper. While the majority of works that used \EPIC{} have focused on action recognition and anticipation, in line with the defined challenges, our dataset lends itself naturally to a variety of less explored tasks. Of these, recent research has explored using \EPIC{} for: video object reasoning and detection~\cite{Baradel_2018_ECCV,Wu_2019_ICCV}, action retrieval~\cite{Wray_2019_ICCV}, visual learning of novel words~\cite{suris2019learning}, unsupervised domain adaptation~\cite{Munro_2020_Arxiv} and learning environmental affordances~\cite{Nagarajan_2020_Arxiv}. Using \EPIC{} for these tasks has only been made possible due to the choices made when collecting this dataset. As opposed to deciding a task then collecting relevant data (or datasets), we opted for capturing natural interactions from which different tasks could be addressed. A few to-be-explored tasks on our dataset, that we are considering for future work are: dense captioning, skill understanding~\cite{Doughty2019Pros} and multi-day routine modelling~\cite{Xu2018}.}

\section*{Data access statement}
Our data is available for download from the Research Data
Repository of University of Bristol at \url{http://dx.doi.org/10.5523/bris.3h91syskeag572hl6tvuovwv4d} along with relevant consent forms.

\ifCLASSOPTIONcaptionsoff
  \newpage
\fi

{\small
\bibliographystyle{IEEEtran}

}

\begin{IEEEbiographynophoto}{Dima Damen}
received PhD (2009) from the University of Leeds. Currently associate professor
in computer vision at University of Bristol, and EPSRC Early Career Fellow, with research interests in the
automatic understanding of object interactions, actions and activities using
static and wearable visual sensors. Supervisor for PhD students Doughty,
Kazakos,
Munro, Price and PhD alumni Moltisanti and Wray. 
Advisor for postdoc researchers Perrett and Wray.
\end{IEEEbiographynophoto}

\begin{IEEEbiographynophoto}{Hazel Doughty}
received MEng (2016) in computer science from the University of Bristol. Currently a fourth year PhD student at the University of Bristol, with research interests in skill understanding from video.
\end{IEEEbiographynophoto}

\begin{IEEEbiographynophoto}{Giovanni Maria Farinella}
received PhD (2008) from University of Catania. Currently associate professor in computer vision and machine learning at the Department of Mathematics and Computer Science, University of Catania, Italy. His research interests lie in the fields of Computer Vision and Machine Learning with focus on Egocentric Perception. Advisor for postdoctoral researcher Furnari.
\end{IEEEbiographynophoto}

\begin{IEEEbiographynophoto}{Sanja Fidler}
received PhD (2010) in computer vision from University of Ljubljana. Currently associate professor at University of Toronto and Vector Institute, Canada CIFAR Artificial Intelligence Chair and director of AI at NVIDIA. Her research interests are in object detection, 3D scene understanding and the interaction between vision and language.
\end{IEEEbiographynophoto}

\begin{IEEEbiographynophoto}{Antonino Furnari}
received PhD (2017) from the University of Catania. Currently postdoctoral researcher of the Department of Mathematics and Computer Science at the University of Catania. His research interests are in computer vision, pattern recognition and machine learning, with focus on First Person (Egocentric) Vision.
\end{IEEEbiographynophoto}

\begin{IEEEbiographynophoto}{Evangelos Kazakos}
received MSc (2017) in computer science from the University of Ioannina. Currently a third year PhD student at the University of Bristol, with research interests in multi-modal fusion for action recognition.
\end{IEEEbiographynophoto}

\begin{IEEEbiographynophoto}{Davide Moltisanti}
received PhD (2019) in computer science from the University of Bristol. Currently a research fellow at Nanyang Tech University, with research interests in levels of supervision for action recognition in video understanding.
\end{IEEEbiographynophoto}

\begin{IEEEbiographynophoto}{Jonathan Munro}
received MEng (2017) in computer science and electrical engineering from the University of Bristol. Currently a third year PhD student at the University of Bristol, with research interests in domain adaptation within action recognition.
\end{IEEEbiographynophoto}

\begin{IEEEbiographynophoto}{Toby Perrett}
recieved PhD (2016) in computer vision from the University of Bristol. Currently a postdoctoral researcher at the University of Bristol, with research interests in transfer and multi-domain learning.
\end{IEEEbiographynophoto}

\begin{IEEEbiographynophoto}{Will Price}
received BSc (2017) in computer science from the University of Bristol. Currently a third year PhD student at the University of Bristol, with research interests in temporal modelling and causality in video.
\end{IEEEbiographynophoto}

\begin{IEEEbiographynophoto}{Michael Wray}
received PhD (2019) in computer science from the University of Bristol. Currently a postdoctoral researcher at the University of Bristol, with research interests in vision and language for fine-grained action recognition and retrieval.
\end{IEEEbiographynophoto}

\newpage
\section*{Supplementary Material} 
\section*{A. Dataset Release}\label{sec:dataset}
\begin{itemize}[leftmargin=*]
\item Dataset sequences, extracted frames and optical flow are available at: \\\textcolor{blue}{\underline{\url{http://dx.doi.org/10.5523/bris.3h91syskeag572hl6tvuovwv4d}}}

\item Annotations, challenge leader-board results and updates and news are available at: \textcolor{blue}{\underline{\url{http://epic-kitchens.github.io}}}
\end{itemize}

\section*{B. Implementation Details}
\revision{This section reports implementation details of the compared baselines for the three challenges in Section 4.}

\subsection*{B.1. Object Recognition Benchmark}
\revision{\textbf{Implementation\hspace{1mm}}
We use the Faster RCNN implementation from the Tensorflow Object Detection API~\cite{objapi2, huang2017speed2}} with a base architecture of ResNet-101~\cite{resnet2} pretrained on MS-COCO~\cite{coco}.

\noindent \revision{\textbf{Training Details\hspace{1mm}}}
Learning rate is initialised to 0.0003 decaying by a factor of 10 after 30000 and 40000 iterations.  We use a per-GPU minibatch size of 4 on 8 NVIDIA P100 GPUs on a single compute node (NVIDIA DGX-1) with distributed training and parameter synchronisation -- i.e. overall minibatch size of 32. As in~\cite{ren2015faster2}, images are rescaled such that their shortest side is 600 pixels and the aspect ratio is maintained. We use a stride of 16 on the last convolution layer for feature extraction and for anchors we use 4 scales of 0.25, 0.5, 1.0 and 2.0; and aspect ratios of 1:1, 1:2 and 2:1. To reduce redundancy, NMS is used with an IoU threshold of 0.7. In training and testing we use 300 RPN proposals.
\subsection*{B.2. Action Recognition Benchmark}
\revision{\textbf{Implementations\hspace{1mm}}
We use the official PyTorch~\cite{pytorch2} implementations of TSN~\cite{tsnpytorch2}, TRN~\cite{trnpytorch2}, and TSM~\cite{tsmpytorch2} pre-trained on ImageNet~\cite{imagenet2}.
We implement 2SCNN as a special case of TSN where the number of segments is 1.
We adapted the code of each baseline to have two output FC layers, one for predicting verbs and the other for nouns.
Our trained models are made available online at \url{https://github.com/epic-kitchens/action-models}.
}

\noindent \revision{\textbf{Modality details\hspace{1mm}}
For RGB networks, the input (per segment) is a single frame and for a flow network it is a stack of 5 $(u,v)$ optical flow pairs (proposed in the two-stream CNN~\cite{simonyan2014twostream2}).
We sample inputs from $n$ segments in training our models.
For TSN, $n = 8$ in training and $n = 25$ in testing.
For TRN and TSM, we set $n = 8$ in both training and testing.}
Optical flow extracted is using the $\tvl$ algorithm~\cite{zach2007duality2} between RGB frames using the formulation $\tvl\left(I_{2t}, I_{\revision{2(t+3)}}\right)$ to eliminate optical flicker.  We have released the computed flow as part of the dataset.
\added{
For the audio stream, we extract $1.28s$ of the raw waveform.
Since there are action segments $< 1.28s$, we extract $1.28$s of audio from the untrimmed video, allowing the audio segment to extend beyond the action boundaries.
We then convert it
to single-channel (via averaging), and resample it to $24$kHz (originally $48$kHz). From that, a log-spectrogram is calculated using an STFT of window length $10$ms, hop length 5ms and $256$ frequency bands. The result is a 2D representation of size $256\times256$.
We replace the size of the global average pooling layer of BN-Inception to $8\times8$, to be compatible with the size of the spectrogram.}

\noindent \revision{\textbf{Training details\hspace{1mm}}}
%, we clip any values larger than 25.
%The optical flow model is trained with a contiguous sequence of 5 optical flow frames as input $X$ formed such that $X_{2t} = V^x_{2t}$ and $X_{2t + 1} = V^y_{2t + 1}$.
\added{
  We train each model on \revision{4} or 8 NVIDIA %P100
  \revision{V100} GPUs (depending on batch size) on a single compute node (NVIDIA DGX-1) with a batch size of 64 (128 for audio models) for 80 epochs using an ImageNet pretrained model for initialisation.
  %The audio model has been trained with a batch size of 128.
}
We do not perform stratification or weighted sampling, allowing the dataset class imbalance to propagate into the minibatch.
\added{
%SGD is used for optimisation with momentum of 0.9.
  \revision{We optimise networks using SGD with momentum 0.9.}
A weight decay of $5\times 10^{-4}$ is applied and gradients are clipped at 20.
We replace the backbone's classification layer with a dropout layer, setting $p = 0.7$.
We train RGB models with an initial learning rate (LR) of 0.01 for ResNet-50 based models and 0.001 for BN-Inception models.
All flow and audio models are trained with a LR of 0.001.
In testing, models are evaluated using 10 crops (center and corner crops as well as their horizontal flips) for each clip.
The scores from these are averaged pre-softmax to produce a single clip-level score.
Fusion results are obtained by averaging the softmaxed scores obtained for each modality.}

\subsection*{B.3. Action Anticipation Benchmark}
\revision{\textbf{Processing Scheme\hspace{1mm}} The methods originally designed for action recognition, such as Two-Stream CNNs (2SCNN), Temporal Segment Networks (TSN), and Temporal Relational Networks (TRN) perform egocentric action anticipation by processing a video observation of length $\tau_a$ sampled $1$ second before the beginning of the action to be anticipated}. DMR has been designed to anticipate actions from single images, whereas ED observes $16$ frames sampled at a $4fps$, which corresponds to an observation time of $\tau_o=4s$.

\noindent \revision{\textbf{Implementations\hspace{1mm}}} We use the official PyTorch implementation provided by the authors in the case of TSN and TRN, whereas 2SCNN is implemented by setting the number of segments to $1$ in TSN. We re-implemented DMR following the details provided in the paper, whereas ED is implemented as in~\cite{gao2017red2}, with the exception that we do not include the reinforcement module, as our task does not consist in discriminating the action from the background as early as possible.

\noindent \revision{\textbf{Training Details\hspace{1mm}}} 2SCNN and TSN are trained for $160$ epochs with a batch size of $64$, using an initial learning rate of $0.001$ that is divided by a factor of $10$ after $80$ epochs. TSN is implemented with $3$ temporal segments. TRN is trained with similar settings, but with a batch size of $32$. DMR is trained in two stages as described in~\cite{vondrick2016anticipating2}. In the first stage, the Deep Multimodal Regressor is trained to predict from static images the representations of frames appearing in one second. The regressor is based on a Batch-Normalized Inception CNN~\cite{ioffe2015batch2} pre-trained on ImageNet, with the addition of fully connected layers with interleaved units as specified in~\cite{vondrick2016anticipating2}. Training is performed for $65$ epochs (which accounts to several weeks on a single GPU) using stochastic gradient descent with momentum equal to $0.9$, a fixed learning rate equal to $0.1$, and and a batch size of $200$. The optimization is performed using the procedure inspired by Expectation-Maximization proposed by the authors. In the second stage, a multilayer perceptron with an hidden layer of $1024$ units and dropout ($p=0.8$) is trained to perform action anticipation using features extracted by the multimodal regressor. The training has been carried out with Stochastic Gradient Descent for $500$ epochs with momentum equal to $0.9$ and a fixed learning rate equal to $0.01$. ED is trained following the procedures outlined in~\cite{gao2017red2}, with an unsupervised pre-training and a supervised fine-tuning. As input features for this model, we use representations extracted using the RGB and Flow branches of a TSN model with three segments trained for action recognition. RGB and Flow predictions have been merged by late fusion with equal weights in the case of 2SCNN, TSN and TRN. DMR relies only on RGB frames, wheres ED relies on early fusion (i.e., it takes as input the concatenation of RGB and Flow representations). All other parameters use the original values provided by the authors.

{\small
\bibliographystyle{IEEEtran}

}

\end{document}